\documentclass[letterpaper]{article} 
\usepackage[preprint]{aaai2027}  
\usepackage[hyphens]{url}  
\usepackage{graphicx} 
\usepackage{natbib}  
\usepackage{caption} 
\usepackage{algorithm}
\usepackage{algorithmic}

\usepackage{newfloat}
\usepackage{listings}
\DeclareCaptionStyle{ruled}{labelfont=normalfont,labelsep=colon,strut=off} 
\floatstyle{ruled}
\newfloat{listing}{tb}{lst}{}
\floatname{listing}{Listing}

\usepackage{enumitem}
\usepackage{booktabs}
\usepackage{tabularx}
\usepackage{multirow}
\usepackage{makecell}
\usepackage[many]{tcolorbox}

\title{Can LLM design high-quality experiments? \\A Comprehensive and Systematic Benchmark on Autonomous Experimental Design.}
\author{
    Zejun Liu\textsuperscript{\rm 1}\equalcontrib, Jian Wu\textsuperscript{\rm 1}\equalcontrib, 
    Ru Peng\textsuperscript{\rm 2}, Yuliang Ji\textsuperscript{\rm 3}, Dongyuan Li\textsuperscript{\rm 4}, 
    Renhe Jiang\textsuperscript{\rm 4}, Yue Zhang\textsuperscript{\rm 1}\corresponding
}
\affiliations{

    \textsuperscript{\rm 1}Westlake University, \textsuperscript{\rm 2}Zhejiang University\\
    \textsuperscript{\rm 3}Nanjing University of Science and Technology, \textsuperscript{\rm 4}University of Tokyo
}

\begin{document}

\maketitle

\begin{abstract}
AI for Research (AI4Research) leverages AI to automate and improve scientific workflows.
While experimental design is a critical stage of the research process,  prior work has focused primarily on code implementation and execution, overlooking
the importance of this stage, and no benchmark exists to evaluate AI's ability to conduct systematic experiment design. 
To bridge this gap, we propose \textsc{SCOPE}, a \textbf{S}cientific \textbf{CO}mprehensive \textbf{P}lanning \textbf{E}valuation Benchmark constructed from 300 high-quality latest papers across 19 research domains from top-tier venues (e.g., ICML, NeurIPS, and ICLR),
evaluating LLMs on two dimensions: High-Level planning completeness (main, ablation, and analysis experiments) and 
Low-Level configuration accuracy and rationality (datasets, baselines, and metrics). Benchmarking reveals three findings:
(1) most LLMs cannot directly design high-quality experiments; (2) all LLMs exhibit a performance bottleneck in low-level configuration; 
and (3) search mode does not improve design quality. Furthermore, to address these challenges, we propose OptED, a novel agentic workflow to 
optimize LLM-based experimental design, that enhances LLM-based experimental planning through stage isolation, tool augmentation, and 
rule-based constraints,  effectively alleviating the configuration bottleneck. \footnote{Code will be made publicly available upon publication.}
\end{abstract}

\section{Introduction}
AI4Research \cite{chen2025ai4research} has emerged as a rapidly advancing interdisciplinary field, embedding AI into the scientific workflow 
to assist or automate activities from scientific knowledge comprehension, literature retrieval and synthesis, to academic 
writing and peer review. Automatic scientific discovery systems have progressed across the research lifecycle: those 
for hypothesis generation and idea mining, such as Spark \cite{sanyal2025spark} and AlphaResearch \cite{yu2025alpharesearch}; 
those for automated code implementation and experiment execution, such as AIDE \cite{jiang2025aide} and AlphaEvolve \cite{novikov2025alphaevolve}; those for data analysis and theoretical 
interpretation, such as NSF-Scify \cite{rao2025nsf}; and end-to-end frameworks that close the loop across the entire pipeline, 
such as AI-Scientist \cite{lu2024ai,yamada2025ai} and AI-Researcher \cite{tang2026ai}.

\begin{figure}[t]
\centering
\includegraphics[width=0.95\columnwidth]{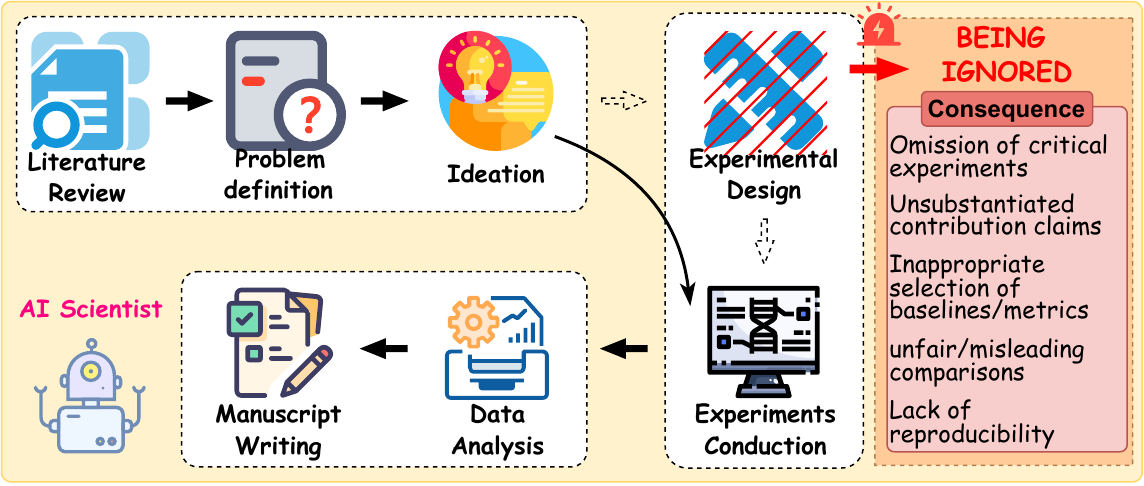} 
\caption{Stages decomposition of scientific discovery lifecycle and the consequence of the lack of experimental design.}
\label{fig1}
\end{figure}

As shown in Figure \ref{fig1}, experimental design occupies a pivotal position in the research lifecycle, bridging methodological conception and empirical validation. It spans High-Level planning, which decides the main experiments, ablation studies, and auxiliary analyses needed to validate research claims, and Low-Level configuration, which selects appropriate datasets, baselines, and metrics. A well-designed scheme directly shapes the credibility of conclusions and reproducibility; conversely, inadequate design triggers cascading failures: critical ablation studies may be omitted, inappropriate baselines or metrics may yield misleading comparisons, or insufficient procedural detail may preclude reproducibility.

Despite these high stakes, existing approaches have largely overlooked this stage, focusing instead on method 
implementation before it and experiment execution after it, reducing what should be an independent, systematic planning 
effort to a mere by-product of the idea-to-code pipeline. Existing benchmarks \cite{kon2025exp,abramovich2025ablationbench,zhao2025abgen} related to experimentation either focus exclusively on code
execution or, in the few cases that address experimental design,
remain confined to isolated sub-tasks such as ablation generation,
falling short of systematic end-to-end assessment.

To address this gap, we introduce \textsc{SCOPE} (Scientific COmprehensive Planning Evaluation Benchmark), featuring a hierarchical task structure that decomposes experimental design into High-Level planning (main experiments, ablation studies, analysis experiments) and Low-Level configuration (datasets, baselines, metrics). Given the research context of a target paper—its task description, related work, and methodology—a model must produce a complete experimental plan covering all six components without access to the original experimental design. Each sub-dimension is independently scored by an LLM-as-Judge on a 0--5 scale against a detailed rubric, yielding a total of 0--30. A redline mechanism zeros scores upon detecting fatal flaws such as source hallucination or metric incompatibility, ensuring that critical failures are not masked by averaging. The dataset comprises 300 tasks across 19 research areas, constructed from papers accepted at top-tier ML venues through a three-stage pipeline of collection, structured extraction, and iterative quality refinement.

Evaluating seven mainstream LLMs under Chain-of-Thought-only (CoT-only) and Chain-of-Thought with search mode (CoT+search) strategies yields three findings: (1) performance varies widely, averaging only 14.81 out of 30 with a 6.07-point gap between the best and worst models; (2) Low-Level configuration accuracy constitutes the core bottleneck, trailing High-Level planning by 2.78 points on average; (3) search access alone fails to improve plan quality---five of seven models show no significant difference, and in some cases it actively degrades High-Level reasoning. Building on these findings, we propose OptED, an agentic workflow that decomposes experimental design into structured stages with tool-augmented atomic operations and explicit behavioral norms. Evaluated on six models, OptED yields consistent improvements, with High-Level and Low-Level scores increasing by 1.6 and 1.2 points on average, while suppressing hallucination-related errors and resolving the search-induced degradation observed in the baseline.

The main contributions of our work are as follows:
\begin{itemize}
    \item We introduce a new dataset and a comprehensive evaluation framework for the systematic assessment of automatically generated experimental plans.
    \item We present a thorough evaluation and analysis of mainstream models in typical operating modes.
    \item We propose an agentic workflow that effectively elevates the quality of experimental design while enhancing the transparency of the overall workflow.
\end{itemize}

\begin{table}[tb]
\centering
\small
\setlength{\tabcolsep}{1.2mm}
\renewcommand{\arraystretch}{1.15}
\begin{tabular}{@{}lllll@{}}
\toprule
Benchmark      & Stage     & Task                                                      & Evaluate                                                            & \begin{tabular}[c]{@{}l@{}}Sys.\\ eval.\end{tabular} \\
\midrule
\makecell[l]{Exp-Bench\\\shortcite{kon2025exp}}      & Exec.     & E2E impl.                                                 & \begin{tabular}[c]{@{}l@{}}LLM-as-Judge,\\ Code valid.\end{tabular} & $\times$                                                  \\[6pt]
\makecell[l]{AblationBench\\\shortcite{abramovich2025ablationbench}}  & Design    & \begin{tabular}[c]{@{}l@{}}Ablation\\ design\end{tabular} & \begin{tabular}[c]{@{}l@{}}LLM-as-Judge(\\ P@k, R@k)\end{tabular}   & $\times$                                                   \\[6pt]
\makecell[l]{Abgen\\\shortcite{zhao2025abgen}}          & Design    & \begin{tabular}[c]{@{}l@{}}Ablation\\ design\end{tabular} & LLM-as-Judge                                                        & $\times$                                                   \\[6pt]
\makecell[l]{SoundnessBench\\\shortcite{ho2026soundnessbench}} & Pre-exec. & \begin{tabular}[c]{@{}l@{}}Proposal\\ judge\end{tabular}  & Macro-F1, R                                                         & $\times$                                                  \\[6pt]
\makecell[l]{AAAR-1.0\\\shortcite{lou2024aaar}}       & Design    & \begin{tabular}[c]{@{}l@{}}Brief\\ design\end{tabular}    & F1, P, R                                                            & $\times$                                                  \\
\midrule
Ours           & Design    & \begin{tabular}[c]{@{}l@{}}Granular\\ design\end{tabular} & LLM-as-Judge                                                        & $\surd$                                                   \\
\bottomrule
\end{tabular}
\caption{Comparison with existing benchmarks related to experimental design. \textit{Stage} denotes the research phase: Exec. (execution), 
Pre-exec. (pre-execution proposal validation), and Design (experimental planning). In \textit{Evaluate}, P and R refer to Precision and Recall, 
respectively. Unlike prior work that focuses on execution, ablation, or brief design, our benchmark is the first to support fine-grained 
experimental design evaluation with system-level assessment.}
\label{benchcomparison}
\end{table}

\section{Related Work}

\subsection{Automatic Systems for Science Discovery}
Prior work has extensively employed LLMs to automate scientific discovery, covering activities that span from literature review to manuscript writing. 
Representative efforts include discovery and synthesize systems such as PaperQA2 \cite{skarlinski2024language}, Pasa \cite{he2025pasa}, Webweaver \cite{li2025webweaver}, 
and Sci-Master \cite{chai2025scimaster}; hypothesis generation and autonomous innovation systems such as Spark \cite{sanyal2025spark}, AutoMind \cite{ou2025automind}, 
FlowPIE \cite{wang2026flowpie}, AlphaResearch \cite{yu2025alpharesearch}, and EvoIdeator \cite{sauter2026evoideator}; code implementation and experiment execution systems 
like Mlr-Copilot \cite{li2024mlr}, AlphaEvolve \cite{novikov2025alphaevolve}, AIDE \cite{jiang2025aide}, and ShinkaEvolve \cite{lange2025shinkaevolve}; data analysis and 
theoretical interpretation systems including NSF-Scify \cite{rao2025nsf}, ML-Master \cite{liu2025ml} and SLDAgent \cite{lin2025can}; Manuscript drafting and reports assembling like SurveyX \cite{liang2025surveyx} and PaperOrchestra\cite{song2026paperorchestra}. 
More recently, a range of end-to-end research frameworks have emerged, including AI-Scientist \cite{lu2024ai,yamada2025ai}, 
AI-Researcher \cite{tang2026ai}, AgentLaboratory \cite{schmidgall2025agent}, AgentRxiv \cite{schmidgall2025agentrxiv}, 
and AutoSDT \cite{li2025autosdt}.

Overall, existing research has largely overlooked the importance of experimental design as an independent phase, 
with dedicated studies remaining notably absent.

\subsection{Benchmarks related to Experiments}
Existing benchmarks largely concentrate on tasks directly tied to coding capabilities, including method 
implementation and experiment execution, as exemplified by MLE-Bench \cite{chan2025mle}, PaperBench \cite{starace2025paperbench}, 
MLRC-Bench \cite{zhang2026mlrc},MLR-Bench \cite{chen2026mlr} and BLADE \cite{gu2024blade}.
In parallel, a subset of benchmarks have begun to take experimental design into consideration, 
including Expbench \cite{kon2025exp}, AbGen \cite{zhao2025abgen}, AblationBench \cite{abramovich2025ablationbench}, AAAR-1.0 \cite{lou2024aaar}, and SoundnessBench \cite{ho2026soundnessbench}.
AbGen and AblationBench both center on ablation study tasks. The core task of ExpBench remains the correct implementation and execution of experiments; 
however, compared to other execution-centric benchmarks, it requires the model to explicitly output a concrete list of parameters and code instructions for each 
individual experiment before coding, thereby incorporating an initial consideration of experimental design into the evaluation. 
SoundnessBench assesses whether models can accurately identify the soundness of research proposals, revealing that their judgment is neither accurate nor stable. 
AAAR-1.0 treats experimental design as a full, independent stage for evaluation, yet its assessment is confined to a brief list of experimental proposals with simple 
matching as the evaluation criterion. Although these benchmarks signal a growing recognition of the importance of experimental design, they have yet to deliver 
a systematic and comprehensive assessment of this phase.

In contrast, our benchmark is purpose-built to fill this gap. It introduces a multi-dimensional evaluation framework that decomposes experimental design into fine-grained, 
independently rated sub-dimensions, thereby enabling a thorough assessment of LLMs' ability to independently generate complete experimental plans. 
Table \ref{benchcomparison} compares our benchmark with existing relevant benchmarks.

\section{\textsc{SCOPE}}

We introduce \textsc{SCOPE}, a benchmark of 300 papers across 19 research domains that formalizes experimental design as a hierarchical structure with six fine-grained evaluation dimensions, assessed via rubric-based LLM-as-Judge augmented with a redline mechanism. Built through a three-stage pipeline of collection, structured extraction, and iterative quality refinement, the benchmark evaluates seven mainstream LLMs under two prompting strategies, revealing systematic bottlenecks that motivate targeted interventions.

\begin{figure*}[t]
\centering
\includegraphics[width=2.1\columnwidth]{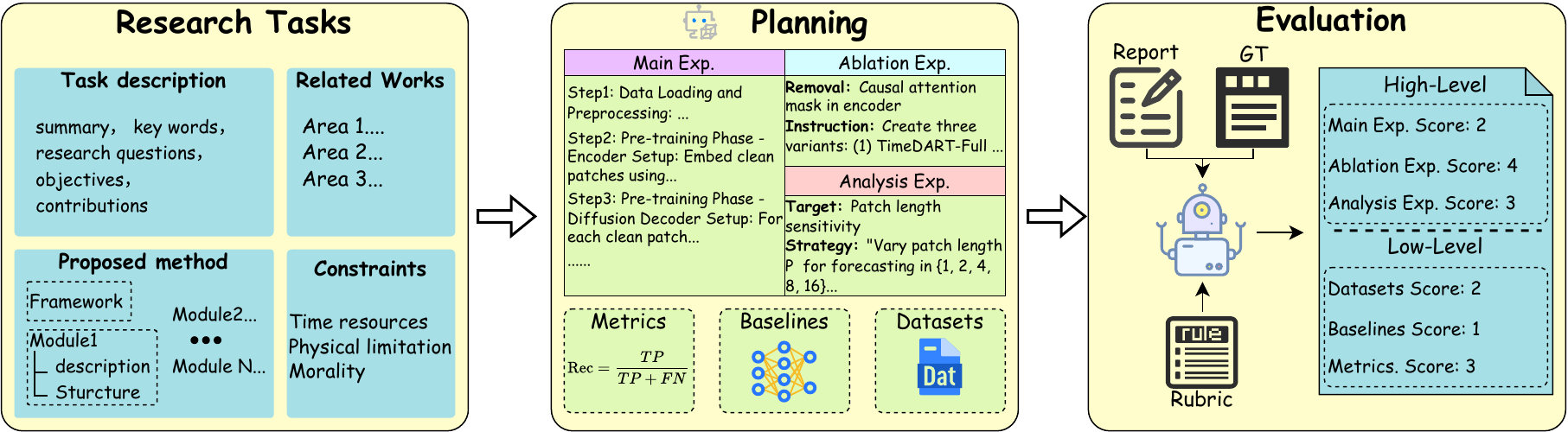} 
\caption{Evaluation pipeline of \textsc{SCOPE}. Given a research task as input, an LLM produces a complete experimental plan spanning six sub-dimensions. The generated plan is then scored against the ground-truth design via rubric-based LLM-as-Judge on a 0--5 scale per dimension, yielding a total of 0--30. A redline mechanism zeros any sub-dimension with fatal flaws.}
\label{evaluate}
\end{figure*}

\subsection{Task Definition}
Experimental design inherently involves two cognitively distinct activities: determining which experiments to 
conduct to validate research claims, and selecting which concrete resources to execute them. We therefore define a standardized 
hierarchical structure comprising two complementary dimensions. The \textbf{High-Level dimension} encompasses three experiment categories: 
main experiments, which construct a complete validation chain addressing the research questions; ablation studies, which isolate and 
verify the independent contribution of each methodological component; and analysis experiments, which provide supplementary insights 
such as hyperparameter sensitivity and computational efficiency. The \textbf{Low-Level dimension} encompasses three resource categories: datasets 
(with sources, splits, and compositions), baseline methods (with sources and performance characteristics), and evaluation metrics 
(both primary and auxiliary). This hierarchical structure transforms experimental design from open-ended free-text generation into 
verifiable structured outputs, enabling systematic evaluation.

Derived from the standardized structure above, we assess generated experimental plans along six sub-dimensions, each corresponding to one component of the hierarchy.
The High-Level sub-dimensions evaluate the scientific soundness and structural completeness of main experiments, ablation studies, and
analysis experiments; the Low-Level sub-dimensions evaluate the accuracy and appropriateness of selected datasets, baselines, and metrics.
Each sub-dimension is independently scored by an LLM-as-Judge on a 0--5 scale against a detailed rubric with explicit per-tier criteria.
The six sub-dimensions sum to a total score of 0--30, with High-Level and Low-Level each contributing 15 points, facilitating clear
identification of model strengths and weaknesses. We further introduce a redline mechanism: when fatal flaws are detected---
fabricating non-existent datasets or baselines (source hallucination), selecting metrics fundamentally misaligned with the task objective
(metric incompatibility), or explicitly violating given constraints (constraint violation)---the corresponding sub-dimension receives a
score of zero regardless of other performance, preventing averaging from masking critical failures and safeguarding the credibility
of evaluation results.

\subsection{Dataset Construction}
The dataset is constructed through three stages: data collection, structured task extraction,
and quality refinement through iterative evaluation; full specifications are provided in Appendix~1.1. 

\textbf{Data Collection:} We collect papers from top-tier venues (ICML, ICLR, NeurIPS) and extract GitHub repository links
from paper PDFs, using GitHub stars and forks as proxies for community recognition and reproducibility.
To ensure that selected papers possess sufficient technical depth and real-world validation for constructing
high-quality experimental design tasks, we rank papers by a composite impact score:

$$R_i = \frac{1}{2}\left[\frac{\ln(1+S_i) - \mu_{\ln S}}{\sigma_{\ln S}} + \frac{\ln(1+F_i) - \mu_{\ln F}}{\sigma_{\ln F}}\right]$$

where $R_i$ is the composite score; $S_i$ and $F_i$ denote GitHub stars and forks; $\mu_{\ln S}$, $\sigma_{\ln S}$, $\mu_{\ln F}$, and $\sigma_{\ln F}$ are the means and standard deviations of the log-transformed values across all papers.
The top 300 papers spanning 19 research areas are selected.

\textbf{Data Curation:} We transform papers into standardized structured representations through 
a two-phase strategy. First, a global understanding phase performs summary-style reading to capture the paper's overall structure, 
core contributions, methodological architecture, and experimental logic, forming a high-level semantic comprehension. Second, a deep 
extraction phase performs fine-grained, section-by-section alignment with the original paper, extracting each component—task description, 
methodology modules, experimental design, datasets, baselines, metrics, constraints, and ambiguities—according to a predefined schema. 
Within methodology, we perform modular decomposition where each module specifies its function, architecture, key formulas, and 
input/output specifications. The experimental design is structured to include main experiment pipelines, ablation studies targeting 
specific modules, and analysis experiments. This summary-then-alignment strategy ensures both holistic coherence and fine-grained accuracy.

\textbf{Quality Refinement:} To ensure data quality, we implement a self-verification and external revision loop. 
Each extracted task first undergoes self-verification, where the model cross-checks the extraction against the original paper 
to identify missing information and hallucinations. Subsequently, an external evaluator scores the task across six 
dimensions—task integrity, related work, methodology, experimental design, configuration accuracy, and other essentials 
(constraints and ambiguities)—against a detailed rubric. Based on the evaluation report, targeted revisions are performed: 
missing details are supplemented by re-examining the paper, hallucinations are corrected against the source text, and vague 
descriptions are refined to ensure executability. This multi-round extraction-evaluation-revision pipeline ensures that each 
task meets rigorous quality standards.

\begin{table*}[tb]
\centering

\setlength{\extrarowheight}{2pt}

\setlength{\aboverulesep}{1pt}
\setlength{\belowrulesep}{1pt}

\begin{tabular}{lccccccccc}
\toprule
\multirow{2}{*}{Model} & \multirow{2}{*}{Strategy} & \multirow{2}{*}{RL-rate} & \multicolumn{3}{c}{High-Level} & \multicolumn{3}{c}{Low-Level} & \multirow{2}{*}{T} \\
\cmidrule(lr){4-6}\cmidrule(lr){7-9}
& & & M & Ab. & An. & D & B & M & \\
\midrule
\multirow{2}{*}{GPT-5.2} & CoT-only  & \underline{1.00}  & \textbf{3.91} & 3.60 & 3.92$^*$ & 2.18$^*$ & 2.15 & 2.45 & 18.22$^*$ \\
\cline{2-10}
& CoT+search & \textbf{0.67} & \underline{3.43} & 2.93 & 3.10 & \textbf{2.39} & \textbf{2.29} & \textbf{2.62} & 16.77 \\
\midrule
\multirow{2}{*}{Claude 4.5 Sonnet} & CoT-only & 2.00$^*$ & 3.19 & \textbf{3.99} & \textbf{4.62} & 2.01 & \underline{2.27} & \underline{2.54} & \textbf{18.62}\\
\cline{2-10}
& CoT+search & 5.33 & 3.30$^*$ & \underline{3.79} & \underline{4.46} & 2.09 & 2.24$^*$ & 2.51$^*$ & \underline{18.38} \\
\midrule
\multirow{2}{*}{Gemini 3 Pro} & CoT-only & 3.33 & 2.47 & 2.47 & 2.24 & 1.85 & 1.72 & 1.91 & 12.65 \\
\cline{2-10}
& CoT+search & 2.67 & 2.45 & 2.45 & 2.25 & 1.85 & 1.71 & 1.95 & 12.66 \\
\midrule
\multirow{2}{*}{Grok-4} & CoT-only & 5.33 & 2.18 & 2.44 & 2.68 & 1.80 & 1.59 & 1.86 & 12.55 \\
\cline{2-10}
& CoT+search & 4.33 & 2.15 & 2.60 & 3.08 & 1.69 & 1.67 & 1.95 & 13.12 \\
\midrule
\multirow{2}{*}{DeepSeek-V3.2} & CoT-only & 7.67 & 2.33 & 2.75 & 3.13 & 1.76 & 1.94 & 2.01 & 13.92 \\
\cline{2-10}
& CoT+search & 14.00 & 2.45 & 2.79 & 3.17 & 1.63 & 1.80 & 1.92 & 13.76 \\
\midrule
\multirow{2}{*}{Qwen3-Max} & CoT-only & 8.67 & 2.40 & 2.61 & 2.79 & 1.64 & 1.75 & 1.86 & 13.05 \\
\cline{2-10}
& CoT+search & 9.33 & 2.40 & 2.55 & 2.80 & 1.66 & 1.62 & 1.85 & 12.88 \\
\midrule
\multirow{2}{*}{Kimi-k2} & CoT-only & 13.33 & 2.87 & 3.32$^*$ & 3.74 & 1.68 & 1.89 & 2.12 & 15.62 \\
\cline{2-10}
& CoT+search & 9.33 & 2.88 & 3.24 & 3.73 & 1.69 & 1.99 & 2.08 & 15.60 \\
\midrule
Qwen-DeepResearch & -- & 9.33 & 2.76 & 2.47 & 2.66 & 1.96 & 1.93 & 2.15 & 13.93 \\
\midrule
Grok-DeepSearch & -- & \textbf{0.67} & 3.07 & 2.71 & 2.93 & \underline{2.23} & 2.12 & 2.31 & 15.39 \\
\bottomrule
\end{tabular}
\caption{Benchmark results of mainstream LLMs under two prompting strategies. RL-rate denotes the redline rate (\%), and T denotes the total score. Sub-dimension abbreviations: M (Main Experiment), Ab (Ablation), An (Analysis), D (Datasets), B (Baselines), M (Metrics). Within each column, \textbf{bold} marks the best value, \underline{underline} the second best, and $^*$ the third best (lower is better for RL-rate).}
\label{benchmark result}
\end{table*}

\subsection{Experiments}
\textbf{Experimental Setup.} We evaluate seven mainstream LLMs: GPT-5.2, Claude 4.5 Sonnet, DeepSeek-V3.2, Gemini 3 Pro, Kimi-K2, Qwen3-Max, and Grok-4. 
Each model is tested under two prompting strategies: CoT-only, where the model performs chain-of-thought reasoning to directly produce 
an experimental plan; and CoT+Search , where the model is additionally provided with web search capability and instructed to invoke 
search proactively during reasoning when needed. We also evaluate two deep research models, Qwen DeepResearch and Grok DeepSearch, which perform 
autonomous multi-step information retrieval and synthesis.

\textbf{Evaluation Protocol.} Figure \ref{evaluate} shows the overview of evaluation process. To ensure validity and fairness, all models are explicitly instructed to output experimental plans following a predefined standardized structure,
which includes: listing selected datasets and baselines with full metadata, designing detailed procedures and resource configurations for main
experiments, ablation studies, and analysis experiments, and providing rationale for each decision. In the Think+Search condition, models are
granted web search access and informed that they may invoke search proactively during reasoning; however, they are strictly prohibited from
directly searching for or accessing the original papers or the proposed methods, and are limited to publicly available resources published before
certain time, simulating the real-world scenario where existing experimental designs are unknown. 

All generated plans are evaluated by GPT-5.2 as judge against the ground-truth experimental design, using a detailed rubric on six sub-dimensions, 
each scored on a 0--5 scale, yielding a total score of 0--30. We report mean scores of each sub-dimension and total score, and redline rates. Full evaluation rubrics and prompts are provided in Appendix~1.2.

\subsubsection{Results and Analysis.}
Table \ref{benchmark result} reports the benchmark results, from which we distill four key observations that characterize 
the current capabilities and limitations of LLMs in experimental design.

\textbf{Observation 1: Model performance varies widely, and most LLMs cannot directly produce high-quality experimental plans.} 

As shown in the benchmark results, Claude 4.5 Sonnet achieves the highest total score of 18.62, while 
Grok-4 scores only 12.55, yielding a gap of 6.07 points. The average across all evaluated models is merely 
14.81/30. Only GPT-5.2 and Claude exceed 18 points, while the remaining models cluster within 12--16 points, 
indicating that mainstream LLMs are generally insufficient for autonomous experimental design. 
Even the top-scoring model falls more than 11 points short of the maximum, underscoring the difficulty of
this task, which demands competence across multiple sub-dimensions that no single model simultaneously achieves.

\textbf{Observation 2: All models exhibit a systematic bottleneck in Low-Level configuration.} 

Across all models, High-Level sub-dimensions (Main Experiment, Ablation Studies, Analysis Experiments) 
consistently and significantly outperform Low-Level sub-dimensions (Datasets, Baselines, Metrics). 
Taking Claude 4.5 Sonnet as an example, its Analysis Experiments score reaches 4.62, while its Datasets score is only 
2.01, a more than twofold gap. No Low-Level sub-dimension exceeds 3 out of 5 for any model, whereas 
High-Level dimensions frequently surpass 3.5. This pattern is consistently reproduced across all seven 
models, confirming that configuration accuracy---rather than planning logic---constitutes the core 
bottleneck, consistent with the mean High-Level/Low-Level gap of 2.78 points noted in the introduction. This persistent asymmetry suggests that Low-Level configuration must be treated as a distinct, externally grounded process rather than an appendage to High-Level planning.

\textbf{Observation 3: Access to search alone does not improve experimental design quality and may introduce additional configuration errors.} 

Comparing the two strategies, five of the seven models show no statistically significant difference in
total score. GPT-5.2, however, exhibits statistically significant degradation under search:
total score drops from 18.22 to 16.77 (--1.45), with High-Level sub-dimensions declining by 0.48--0.82 points each while Low-Level dimensions see only marginal gains of 0.14--0.21 points.
This pattern---search trading off High-Level planning for marginal Low-Level gains---manifests to varying degrees across other models.
Furthermore, DeepSeek-V3.2's redline rate nearly doubles from 7.67\% to 14.00\% under search, indicating that additional information from search increases hallucination risk for certain models.
Simply granting search access not only fails to address the core challenges but may actively disrupt the model's reasoning chain by introducing unfiltered external information. The core issue is not access to information but how it is integrated: unstructured retrieval competes with, rather than complements, internal reasoning.

\textbf{Observation 4: Deep research models improve over their base counterparts but do not resolve the core bottleneck or close the gap with top models.} 

Grok DeepSearch achieves 15.39, substantially improving over its base model (+2.84), with particularly pronounced Low-Level gains (Datasets +0.43, Baselines +0.53) and a redline rate of only 0.67\%, the lowest among all models. 
Qwen DeepResearch similarly outperforms its base model across all dimensions. These gains stem from deep research models' 
multi-step autonomous retrieval and information synthesis capability, which enables progressive acquisition, filtering, and 
integration of external information over longer reasoning chains. However, deep research does not alter the systemic bottleneck: 
Grok DeepSearch's Low-Level scores (Datasets 2.23, Baselines 2.12, Metrics 2.31) remain substantially below its High-Level scores 
(Main 3.07, Analysis 2.93), and none exceed the 3-point threshold. More critically, Grok DeepSearch still trails far behind 
Claude 4.5 Sonnet and GPT-5.2, demonstrating that enhanced retrieval depth alone cannot bridge the fundamental gap in planning 
and reasoning capability.

Collectively, these observations point to two design imperatives: Low-Level configuration must be handled independently from planning, and external knowledge must be integrated through structured workflows rather than unstructured retrieval---both of which directly motivate OptED.

\section{OptED}

Motivated by the \textsc{SCOPE} findings---that Low-Level configuration accuracy constitutes the core bottleneck and that naive search integration can degrade High-Level planning---we propose OptED, an agentic workflow that reframes experimental design from single-pass generation into a structured, tool-augmented process through stage isolation, atomic tool operations, and explicit behavioral norms. We evaluate OptED across six models and isolate the contribution of each component through ablation studies.

\subsection{Framework}
OptED is a three-stage agentic workflow that reframes experimental design from open-ended text generation into a controlled, 
structured working process. 

\textbf{Stage isolation.} Experimental design inherently involves two cognitively distinct activities: Low-Level resource selection and High-Level protocol planning.
Mixing them in a single context forces the model to repeatedly switch attention between the two, leading to cross-stage
hallucination and information degradation. We therefore decouple the workflow into three linearly connected but contextually
independent stages. The \textbf{Configuration Phase} independently handles Low-Level resource discovery, evaluation, and selection. The \textbf{Protocol Phase} builds upon the previous stage's output to construct the complete High-Level scheme of main
experiments, ablation studies, and analysis experiments. The \textbf{Reporting Phase} assembles the structured results into the
final experimental plan. Stage isolation ensures that each stage operates with a dedicated tool set and independent context,
preventing cross-stage interference; stage boundaries serve as natural checkpoints, making the workflow transparent, traceable,
and amenable to backtracking.

\textbf{Tool-Augmented Think-Act-Observe Loop.} Rather than generating an experimental plan in a single pass, the agent operates under the Think-Act-Observe paradigm: 
it first analyzes the current state and determines the next operation (Think), invokes the corresponding tool with structured arguments (Act), 
and receives the tool output as feedback for the next round (Observe). This iterative loop decomposes open-ended experimental design into an 
explicit, traceable multi-step workflow, where each round produces a verifiable intermediate result.

The tool layer is deeply coupled with this paradigm. Its design principle is to atomize complex experimental design actions into structured, 
verifiable tool interfaces, thereby converting free-text generation into executable, auditable structured operations. It comprises two functional 
categories. \textbf{Search and browsing tools} support multi-source retrieval, full-text extraction, and structured paper comprehension, 
following a hypothesize-then-verify pattern to avoid aimless exploration. \textbf{Content editing tools} encapsulate CRUD operations on datasets, 
baselines, and experiment protocols as atomic primitives, with each invocation mandatorily recording the operation content, state change, and 
decision rationale. The specific editing tools provisioned to each phase differ according to its task type and output requirements: the 
Configuration Phase operates on datasets and baselines, while the Protocol Phase operates on experiment steps, ablation studies, 
and analysis experiments.

\textbf{Behavioral Norms.} Tools equip the model with the capability to act, but the model still requires guidance to wield these capabilities correctly. 
We therefore embed explicit behavioral norms and checklists within each stage, defining standard operating procedures and decision 
criteria. In the Configuration Phase, the norms span the full pipeline from task assessment, internal knowledge recall, and 
multi-source discovery to principle-based candidate screening, specifying the scientific standards that dataset and baseline 
selection should satisfy. In the Protocol Phase, the norms constrain the design requirements for each experiment category—ablation 
studies must follow the single-variable principle, and analysis experiments should provide insights beyond what the main results 
table reveals. The checklist triggers a global self-check at the end of each stage, ensuring no critical step is overlooked. 
Together, these norms converge the LLM's open-ended behavior into a structured, reproducible research operation mode. 
Full implementation details are provided in Appendix~2.

\subsection{Evaluation}

\begin{figure*}[t]
\centering
\includegraphics[width=2\columnwidth]{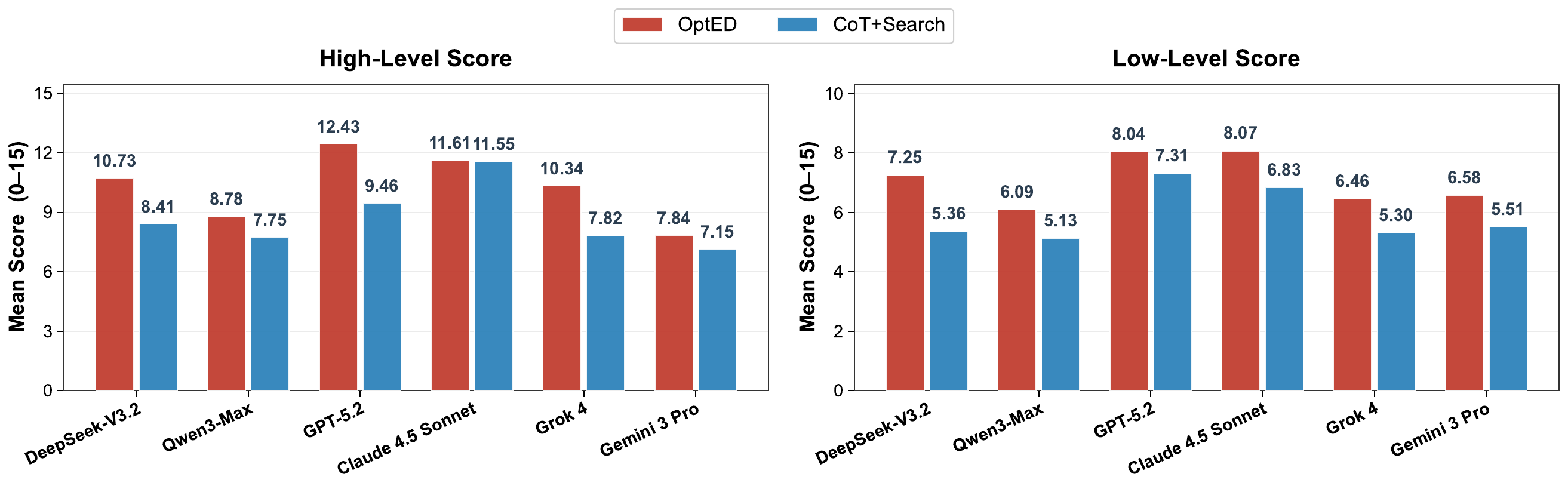} 
\caption{Scores comparison between OptED and the CoT+Search baseline across six models. OptED yields consistent gains across all models in both High-Level and Low-Level dimensions, with DeepSeek-V3.2 (+4.22) and Grok-4 (+3.67) benefiting most. GPT-5.2 achieves the highest absolute score (20.47), fully recovering the search-induced degradation observed in the baseline.}
\label{comparison}
\end{figure*}

\begin{figure}[t]
\centering
\includegraphics[width=1\columnwidth]{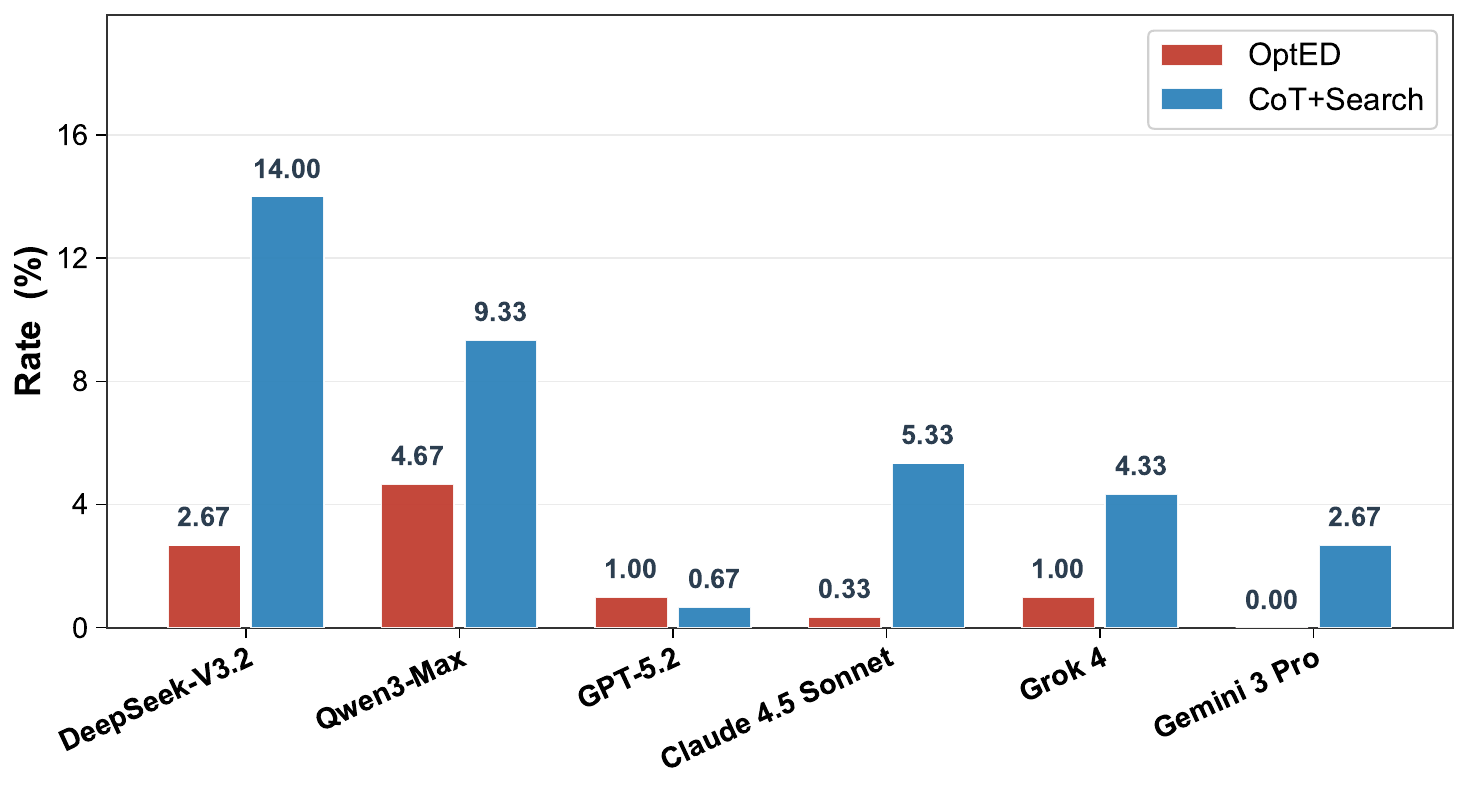} 
\caption{RL-rate comparison between OptED and the CoT+Search baseline across six models. Redline rates drop substantially under OptED across all models, with the strongest reduction from 14.00\% to 2.67\%, and Gemini 3 Pro achieving zero errors.}
\label{RL}
\end{figure}

We evaluate OptED on six models: GPT-5.2, Claude 4.5 Sonnet, DeepSeek-V3.2, Grok-4, Gemini 3 Pro, and Qwen3-Max.
Since OptED integrates search within a structured workflow, we compare against the CoT+Search baseline to directly test whether the 
framework addresses the issues identified in Observation 3. Figure \ref{comparison} and Figure \ref{RL} report the results.

OptED yields consistent gains across all six models. DeepSeek-V3.2 (+4.22, +30.7\%) and Grok-4 (+3.67, +28.0\%) benefit most, confirming that models with the weakest configuration stability under CoT+Search---the pattern diagnosed in Observation 2---gain the most from structured decomposition. GPT-5.2 and Claude 4.5 Sonnet achieve the highest absolute scores of 20.47 and 19.68, respectively, while Gemini 3 Pro (+1.76) and Qwen3-Max (+1.98) obtain steady improvements. Gains span both High-Level and Low-Level dimensions across all models, confirming generality.

Notably, GPT-5.2 fully reverses the search-induced degradation identified in Observation 3: from a CoT+Search low of 16.77, OptED pushes its score to 20.47, exceeding its CoT-only peak (18.22) by 2.25 points. This directly validates that decoupling search from planning into distinct phases preserves reasoning coherence while harnessing external knowledge, resolving the retrieval-reasoning interference that the benchmark analysis exposed.

OptED also addresses the hallucination issue surfaced in Observation 3: redline rates drop from 14.00\%
to 2.67\% for DeepSeek-V3.2, to 0.33\% for Claude 4.5 Sonnet, and to zero for Gemini 3 Pro. The atomic tool interfaces with mandatory rationale recording prevent the source fabrication and metric incompatibility that unfiltered search had introduced. GPT-5.2 maintains its already low 1.00\%, reflecting a floor effect for models with intrinsically high factuality.

Despite these gains, the High-Level/Low-Level gap persists across all models: GPT-5.2 scores 12.43 in planning against 8.04 in configuration, and similar disparities hold for the remaining models. This confirms that while structured workflows effectively organize the design process, grounding plans in precise resource choices remains beyond the reach of workflow optimization alone---reinforcing the Observation 2 finding that Low-Level configuration is the deeper, harder challenge.

\subsection{Ablation Study}
We conduct ablation experiments on GPT-5.2, DeepSeek-V3.2, and Qwen3-Max to isolate the contribution of each component. 
Three configurations are compared: full OptED; w/o norms, which retains stage isolation and the Think-Act-Observe loop 
but removes behavioral norms; and stage isolation only, which retains the three-stage structure and search capability but 
removes both norms and the T-A-O loop, i.e., the model generates directly within each stage without multi-step atomic operations. Table \ref{ablation} reports the results.

\begin{table}[t]
\centering
\small
\begin{tabular}{lccccc}
\toprule
Model & Config & RL-rate & HL & LL & T \\
\midrule
\multirow{3}{*}{GPT} & full OptED & 1.00 & 12.43 & 8.04 & 20.47 \\
& w/o norms & 1.00 & 12.32 & 7.98 & 20.31 \\
& stage only & 0.67 & 10.83 & 7.68 & \textbf{18.51} \\
\midrule
\multirow{3}{*}{DS} & full OptED & 2.67 & 10.73 & 7.25 & 17.98 \\
& w/o norms & 2.33 & 10.33 & 6.95 & 17.28 \\
& stage only & 4.33 & 9.19 & 5.76 & \textbf{14.95} \\
\midrule
\multirow{3}{*}{Qwen} & full OptED & 4.67 & 8.78 & 6.09 & 14.86 \\
& w/o norms & 4.33 & 8.64 & 5.82 & 14.46 \\
& stage only & 5.33 & 8.19 & 5.54 & \textbf{13.73} \\
\bottomrule
\end{tabular}
\caption{Ablation results across three models. GPT, DS (DeepSeek), and Qwen denote GPT-5.2, DeepSeek-V3.2, and Qwen3-Max, respectively. HL and LL denote the mean High-Level and Low-Level total scores.}
\label{ablation}
\end{table}

Stage isolation alone yields consistent gains over the CoT+Search baseline across all three models, confirming the 
generality of structural decoupling even without multi-step atomic operations. Introducing the T-A-O loop (w/o norms) 
brings substantial further improvements: DeepSeek-V3.2 gains most (+2.33), followed by GPT-5.2 (+1.80) and Qwen3-Max (+0.73), 
demonstrating the value of converting free-form generation into explicit, traceable multi-step workflows. The contribution of 
behavioral norms varies with model capability: DeepSeek-V3.2 (+0.70) and Qwen3-Max (+0.40) benefit clearly, whereas GPT-5.2 
improves only marginally from 20.31 to 20.47 (+0.16). This suggests that GPT-5.2 already possesses sufficiently strong 
internal reasoning discipline, leaving limited room for external norms to add value; for models with weaker intrinsic 
planning consistency, norms serve as an effective compensatory mechanism. Regarding redline rates, stage isolation alone 
already reduces DeepSeek-V3.2 and Qwen3-Max from their baseline rates of 14.00\% and 9.33\% to 4.35\% and 5.33\%, indicating 
that structural decoupling itself effectively suppresses cross-stage hallucination; subsequent components further compress 
these to 2.67\% and 4.67\%, respectively.

\section{Conclusion}
This work systematically investigated LLM-based autonomous experimental design. We introduced \textsc{SCOPE}, 
a benchmark that identified Low-Level configuration accuracy as the core bottleneck and revealed that naive search degrades 
planning quality. We then proposed OptED, a stage-isolated workflow that resolved this degradation and yielded consistent 
gains across all models---reciprocally validating our diagnosis. The persistent gap between planning and configuration, 
however, remains an open challenge for future work.

\bibliography{aaai2027}


\appendix
\section*{Acknowledgment of LLM Assistance}

We acknowledge the use of large language models to assist with manuscript preparation. Specifically, LLMs were employed solely for language refinement tasks, including grammatical corrections, vocabulary enhancement, and sentence structure improvements. All scientific content, including research methodology, experimental design, data analysis, theoretical developments, and technical conclusions, represents the original intellectual contribution of the authors. The LLMs did not generate any substantive scientific content, interpretations, or novel insights presented in this work.

\section{Benchmark Specification}

This section provides detailed specifications of the \textsc{SCOPE} benchmark, complementing the concise descriptions in the main paper. We first present the dataset construction methodology, including the complete input/output schema, domain coverage, and quality control mechanisms (Section~1.1). We then elaborate on the evaluation protocol, covering the full scoring rubric, the redline mechanism, and the statistical analysis framework (Section~1.2).

\subsection{Dataset Specification and Construction}

The \textsc{SCOPE} dataset consists of 300 high-quality research tasks derived from papers accepted at ICML, ICLR, and NeurIPS, 
curated to evaluate an LLM's ability to independently design experimental plans from methodological understanding alone. 
Each task provides a model with the research context of a target paper---its task description, related work, and methodology---
and requires the model to produce a complete experimental plan covering both high-level experiment types and low-level 
resource configurations. The ground truth for each task is the original paper's full experimental design, structured into 
a standardized hierarchical format. This formulation mirrors real-world research practice, where experiments must be designed 
solely from methodological understanding without access to experimental outcomes.

\subsubsection{Overview and Key Statistics}

\begin{tcolorbox}[
	title={Box 1: \textsc{SCOPE} Dataset at a Glance},
	colback=blue!3!white,
	colframe=blue!30!black,
	colbacktitle=blue!15!white,
	coltitle=black,
	fonttitle=\bfseries,
	arc=3mm,
	boxrule=0.8pt,
	breakable
]
\small
\renewcommand{\arraystretch}{1.12}
\begin{tabularx}{\linewidth}{@{}>{\bfseries}l@{\quad}>{\raggedright\arraybackslash}X@{}}
\toprule
\multicolumn{2}{@{}l@{}}{\textbf{Dataset Statistics}} \\
\midrule
Total tasks          & 300 \\
Research domains     & 19 \\
Source venues        & ICML, ICLR, NeurIPS  \\
Initial candidate pool & $>$5{,}200 papers \\
Avg.\ modules per method & 4.2 \\
Avg.\ baselines per task  & 6.8 \\
Avg.\ datasets per task   & 3.1 \\
Quality threshold    & $\geq$35/50 (70\%, ``Good'' grade or above) \\
\bottomrule
\end{tabularx}
\end{tcolorbox}

\medskip

The 300 tasks were selected from an initial pool of over 5,200 papers collected from the three venues. Each paper in the candidate pool underwent a rigorous multi-stage pipeline of collection, enrichment, impact-based ranking, structured extraction, quality assessment, and targeted revision. Only tasks that passed a minimum quality threshold of 35 out of 50 points (70\%, corresponding to the ``Good'' grade or above) in the quality assessment stage were retained, ensuring that every task meets a baseline standard of completeness and accuracy before entering the benchmark.

\subsubsection{Domain Coverage}

The dataset spans 19 standardized research domains, covering the breadth of contemporary machine learning research. Table~\ref{tab:domains} lists the complete domain taxonomy along with representative subtopics.

\begin{table}[htbp]
\centering
\footnotesize
\setlength{\tabcolsep}{1mm}
\renewcommand{\arraystretch}{1.05}
\begin{tabularx}{\columnwidth}{@{}c@{\hskip 1.5mm}>{\raggedright\arraybackslash}p{3.5cm}>{\raggedright\arraybackslash}X@{}}
\toprule
\textbf{ID} & \textbf{Domain} & \textbf{Representative Subtopics} \\
\midrule
0 & Vision, Audio, Language Apps. & image recognition, video, speech, multimodal learning \\
1 & Generative Models & diffusion, GANs, autoregressive, image/video synthesis \\
2 & Robotics, Autonomy, Planning & manipulation, navigation, motion planning, autonomous driving \\
3 & Foundation / Frontier Models (incl.\ LLMs) & LLMs, pre-training, scaling laws, in-context learning \\
4 & Datasets and Benchmarks & benchmark construction, data collection, evaluation protocols \\
5 & Interpretability \& Explainability & feature attribution, concept explanations, mechanistic interpretability \\
6 & Representation Learning & self-supervised, contrastive, multi-view learning \\
7 & Reinforcement Learning & policy optimization, model-based RL, multi-agent RL \\
8 & Alignment, Fairness, Safety, Privacy & RLHF, bias mitigation, adversarial robustness, differential privacy \\
9 & Other Topics in ML & emerging paradigms, interdisciplinary methods \\
10 & Optimization & convex/non-convex, stochastic, distributed training \\
11 & Infrastructure, Systems, Hardware & ML compilers, distributed systems, hardware-aware algorithms \\
12 & Applications to Physical Sciences & AI for physics, chemistry, biology, drug discovery, protein design \\
13 & Time Series \& Dynamical Systems & forecasting, state-space models, neural ODEs \\
14 & Neuroscience \& Cognitive Science & neural coding, brain-computer interfaces, cognitive models \\
15 & Transfer, Meta, Lifelong Learning & few-shot learning, domain adaptation, continual learning \\
16 & Neurosymbolic \& Hybrid AI & physics-informed NNs, logic reasoning, knowledge integration \\
17 & Learning on Graphs \& Geometries & GNNs, geometric DL, molecular graphs, manifold learning \\
18 & Learning Theory & generalization bounds, PAC learning, optimization landscapes \\
\bottomrule
\end{tabularx}
\caption{The 19 standardized research domains in \textsc{SCOPE}, adapted from the primary area taxonomies of ICML, ICLR, and NeurIPS.}
\label{tab:domains}
\end{table}

Each paper in the dataset is assigned to one primary domain. The domain distribution reflects the relative prevalence of research topics at the source venues, with foundation models, generative models, and multimodal applications being the most represented areas. This diversity ensures that the benchmark evaluates experimental design capability across a broad spectrum of methodological paradigms, rather than being confined to any single subfield.

\subsubsection{Input Structure}

The input to each benchmark task is a self-contained research context that provides the model with all necessary methodological understanding while strictly withholding information about the original paper's experimental choices. It consists of four components, formally defined below.

\begin{tcolorbox}[
	title={Box 2: Input Schema — Research Task Specification},
	colback=blue!3!white,
	colframe=blue!30!black,
	colbacktitle=blue!15!white,
	coltitle=black,
	fonttitle=\bfseries,
	arc=3mm,
	boxrule=0.8pt,
	breakable
]
\small
\textbf{Component 1:} \texttt{task\_description}
\begin{itemize}[nosep,leftmargin=12pt]
	\item \texttt{summary}: A concise overview of the research, covering the problem setting, proposed approach, and key findings.
	\item \texttt{keywords}: A list of technical keywords characterizing the research topic.
	\item \texttt{research\_area}: One or more research domains to which the work belongs.
	\item \texttt{questions}: The specific research questions that the work aims to answer.
	\item \texttt{research\_objective}: The concrete goals that the methodology is designed to achieve.
	\item \texttt{contributions}: The claimed contributions of the work.
\end{itemize}
\medskip

\textbf{Component 2:} \texttt{related\_work} — A curated list of related prior work, organized by topic or methodology family. Each entry identifies the relevant method or study and its relationship to the target paper, providing the contextual background necessary for identifying appropriate baselines.

\medskip

\textbf{Component 3:} \texttt{method} — A fine-grained modular decomposition of the proposed approach:
\begin{itemize}[nosep,leftmargin=12pt]
	\item \texttt{summary}: A high-level description of the proposed method and its overall architecture.
	\item \texttt{Modules}: An ordered list of modules, each specifying:
	\begin{itemize}[nosep,leftmargin=16pt]
		\item \texttt{name}: The module identifier.
		\item \texttt{description}: Its function and role within the overall method.
		\item \texttt{structure}: The internal architecture, listed in sequential order.
		\item \texttt{key formula or operation}: The core mathematical or algorithmic operations (if applicable).
		\item \texttt{input}, \texttt{output}: The data types and semantics at the module boundary.
	\end{itemize}
	\item \texttt{framework}: The complete composition logic describing how all modules are connected to form the full pipeline.
\end{itemize}
\medskip

\textbf{Component 4:} \texttt{constraints} (optional) — Explicit constraints that the experimental design must respect, such as hardware limitations (e.g., maximum GPU memory), computational budgets, data access restrictions, or ethical considerations. These reflect the incomplete information and practical boundary conditions that researchers face in real-world experimental planning.
\end{tcolorbox}

\medskip

Several design decisions are critical to the benchmark's validity. First, \textbf{all information about the original paper's experimental design is excluded} from the input. The model receives no information about which datasets the original authors used, which baselines they compared against, which metrics they reported, or how they structured their experiments. This forces models to reason about experimental choices from first principles rather than reproducing them from memory or textual cues, and it prevents the task from being solved by superficial pattern matching against the input text.

Second, \textbf{methodology decomposition is performed at the module level} rather than as a monolithic description. Each module specifies its function, architecture, key formulas, and I/O interface, enabling a model to determine which components warrant dedicated ablation studies and which configuration choices are architecturally constrained. This granularity is essential for evaluating a model's ability to design targeted ablation experiments that isolate individual methodological contributions.

Third, \textbf{constraints are explicitly enumerated} when present in the original paper. These may include practical constraints such as GPU memory limits that preclude training large-scale models from scratch, or dataset access restrictions that require the use of publicly available alternatives. By making these constraints explicit, the benchmark assesses whether models can adapt their experimental plans to realistic resource limitations rather than proposing infeasible experiments.

\subsubsection{Ground Truth Definition}

The ground truth for each task is the complete experimental design extracted from the original paper, structured into a two-level hierarchy that captures both the scientific reasoning and the operational details of the experimental plan.

\begin{tcolorbox}[
	title={Box 3: Ground Truth Schema — Hierarchical Experimental Design},
	colback=blue!3!white,
	colframe=blue!30!black,
	colbacktitle=blue!15!white,
	coltitle=black,
	fonttitle=\bfseries,
	arc=3mm,
	boxrule=0.8pt,
	breakable
]
\small
\textbf{High-Level: Experiment Types} (what experiments to conduct)

\medskip
\textit{Main Experiments} — The primary validation pipeline addressing the core research questions:
\begin{itemize}[nosep,leftmargin=12pt]
	\item \texttt{steps}: An ordered list of procedural steps, each with a detailed \texttt{requirement} describing the operation, its purpose, and how it contributes to the validation chain.
	\item \texttt{configuration}: Specifies \texttt{chosen\_datasets}, \texttt{chosen\_baselines}, and \texttt{chosen\_metrics} to be used.
\end{itemize}
\medskip

\textit{Ablation Studies} — Targeted experiments isolating individual methodological components:
\begin{itemize}[nosep,leftmargin=12pt]
	\item \texttt{name}: The ablation study identifier describing the targeted component.
	\item \texttt{removal}: The module(s) removed or replaced in this ablation variant.
	\item \texttt{instruction}: Concrete instructions for executing the ablation (e.g., which component to remove, what to replace it with, how to configure the variant).
	\item \texttt{configuration}: Specifies datasets, baselines, and metrics used for this ablation.
\end{itemize}
\medskip

\textit{Analysis Experiments} — Supplementary investigations providing additional insights:
\begin{itemize}[nosep,leftmargin=12pt]
	\item \texttt{target}: The aspect under investigation (e.g., hyperparameter sensitivity, computational efficiency, robustness to distribution shift).
	\item \texttt{strategy}: The analytical approach (e.g., grid search over learning rates, measuring inference latency vs.\ model size).
	\item \texttt{configuration}: Specifies datasets, baselines, and metrics used for this analysis.
\end{itemize}
\medskip

\textbf{Low-Level: Resource Configuration} (which concrete resources to use)

\medskip
\textit{Datasets} — For each dataset used in the experimental design:
\begin{itemize}[nosep,leftmargin=12pt]
	\item \texttt{name}: Standard name with version if applicable.
	\item \texttt{source}: Accessible URL or acquisition method.
	\item \texttt{description}: Overall introduction and purpose.
	\item \texttt{compositions}: Structural components and data types in each.
	\item \texttt{task}: The specific research task each component targets.
	\item \texttt{data\_format}: Feature specifications and sample formats.
	\item \texttt{label\_info}: Label type, class distribution, and balance characteristics.
	\item \texttt{collection}: Data collection and labeling methodology (crucial for assessing potential biases).
	\item \texttt{size\_and\_split}: Total scale and train/validation/test split statistics.
\end{itemize}
\medskip

\textit{Baselines} — For each baseline method used in comparison:
\begin{itemize}[nosep,leftmargin=12pt]
	\item \texttt{name}: Standard method name.
	\item \texttt{source}: Accessibility information (paper title, arXiv ID, or repository URL).
	\item \texttt{time}: Publication year (used to enforce temporal constraints during evaluation).
	\item \texttt{info}: Brief description of the method and its relevance.
	\item \texttt{performance}: Reported performance on relevant datasets (e.g., ``ImageNet: 76.5\% top-1 accuracy'').
\end{itemize}
\medskip

\textit{Metrics} — Evaluation metrics used across experiments:
\begin{itemize}[nosep,leftmargin=12pt]
	\item \texttt{Primary}: The main metric for performance comparison.
	\item \texttt{Secondary}: A list of auxiliary metrics providing complementary perspectives.
\end{itemize}
\medskip

\textit{Implementation Configuration} — Default hyperparameter settings inferred from the paper:
\begin{itemize}[nosep,leftmargin=12pt]
	\item A list of name--default pairs (e.g., \texttt{learning\_rate: 0.001}, \texttt{batch\_size: 64}).
\end{itemize}
\end{tcolorbox}

\medskip

This hierarchical structure serves two purposes. First, it decomposes the inherently complex activity of experimental design into discrete, independently evaluable components, enabling fine-grained assessment that pinpoints specific strengths and weaknesses. Second, it captures the distinction between \textit{knowing what experiments to run} (High-Level) and \textit{knowing how to configure them} (Low-Level)---a distinction that our benchmark results show is empirically significant, as models consistently exhibit a substantial performance gap between these two levels (2.78 points on average across all evaluated models).

The High-Level dimension assesses whether a model can reconstruct the scientific logic of validation: identifying the core experimental pipeline that tests the research claims, isolating individual components through ablation, and probing method behavior through auxiliary analyses. The Low-Level dimension assesses whether a model can operationalize this logic with appropriate, real-world resources: selecting datasets that match the task domain, baselines that represent meaningful points of comparison, and metrics that faithfully capture the intended evaluation criteria.

\subsubsection{Data Construction Pipeline}

\begin{figure*}[t]
\centering
\includegraphics[width=1.8\columnwidth]{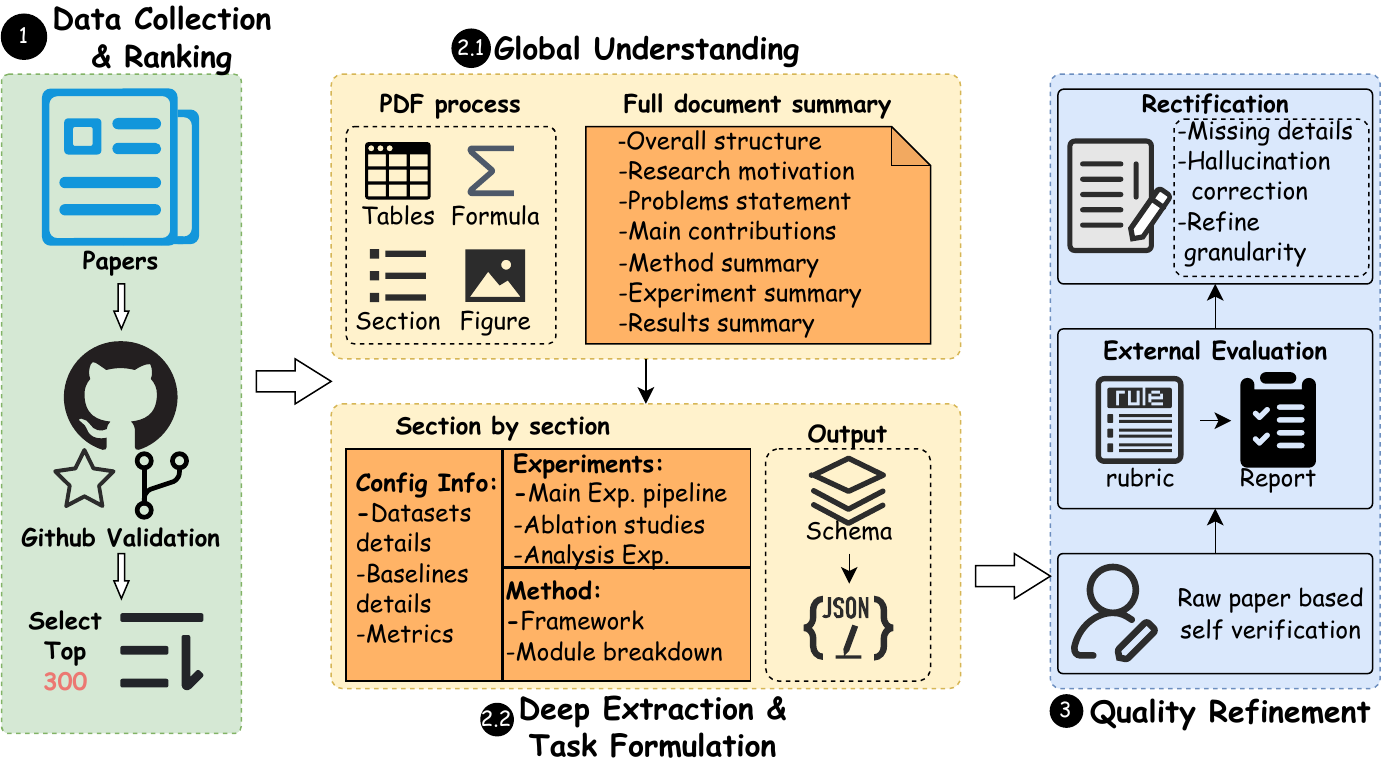}
\caption{The \textsc{SCOPE} dataset construction pipeline. Stage 1 (Collection): papers are selected from top-tier venues and ranked by a composite impact score based on GitHub stars and forks. Stage 2 (Curation): each paper undergoes summary-then-alignment extraction to produce standardized structured representations. Stage 3 (Refinement): a self-verification and external revision loop ensures quality through multi-round evaluation and targeted correction.}
\label{app:dataset construction}
\end{figure*}

The main paper (Section~3.2) outlines the three-stage construction pipeline---collection, curation, and refinement---at a high level. We complement that description here with implementation-level details, the quality assessment rubric, and prompt templates used during extraction and revision. Figure~\ref{app:dataset construction} provides an architectural overview of the pipeline.

\paragraph{PDF Processing and Format Conversion.}
All papers are downloaded from OpenReview as PDFs and converted to Markdown using the Marker library,\footnote{\url{https://github.com/VikParuchuri/marker}} which preserves document structure including section headings, mathematical equations (as \LaTeX{}), tables, and figure captions. Links to external resources---GitHub repositories, project pages, dataset portals---are extracted during this step and used for repository verification and metadata enrichment.

\paragraph{Three-Phase Structured Extraction.}
The extraction of each structured task representation follows a three-phase \textit{summary-then-alignment} workflow executed by a single LLM in a single continuous context. \textbf{Phase 1 (Comprehensive Analysis \& Summarization):} the model first reads the paper in its entirety to construct a holistic cognitive map, producing a hierarchical summary that covers the research overview, keywords, domain, core problem, main contributions, and high-level descriptions of the methodology, experiments, and conclusions. This phase establishes the global semantic context that guides subsequent fine-grained extraction. \textbf{Phase 2 (Deep Extraction \& Task Construction):} with this global understanding in place, the model conducts a granular, section-by-section reading, aligning each extraction target with the corresponding passage in the source text. Specifically, it extracts related work organized by domain; performs a modular decomposition of the methodology (each module specifying function, structure, key formulas/operations, and expected input/output); records datasets with full metadata (source, composition, task, data format, labels, collection, size, splits); distinguishes primary from secondary metrics; reconstructs the experimental design as main experiment pipelines, ablation studies, and analysis experiments; and captures baselines with reported performance, constraints, and any ambiguities in the paper. \textbf{Phase 3 (Verification \& Quality Assurance):} the model re-reads the paper to cross-reference extracted details against the source, corrects misunderstandings, and explicitly records any ambiguous or missing information---such as undefined hyperparameters or contradictory descriptions---that could hinder reproducibility. The full extraction prompt, including the detailed instructions for each phase, is provided in Box~4.

\begin{tcolorbox}[
	title={Box 4: Extraction Prompt of Phase1 and Phase2},
	colback=blue!3!white,
	colframe=blue!30!black,
	colbacktitle=blue!15!white,
	coltitle=black,
	fonttitle=\bfseries,
	arc=3mm,
	boxrule=0.8pt,
	breakable
]
\small
\textbf{Research Task Extraction Guidance}

\textbf{Your Persona:} You are a distinguished senior researcher in the fields of Computer Science and Artificial Intelligence.

\textbf{Your Mission:} You are required to meticulously analyze a complete research paper---including its main text, figures, tables, and appendices. Your goal is to deconstruct the paper and extract fine-grained details to formulate a structured and replicable research task. You must strictly adhere to the following three-step process.

\medskip
\textbf{Execution Steps}

\textbf{Step 1: Holistic Comprehension \& Summary}

Thoroughly read the entire paper to grasp the full scope of the research. You must understand the paper's structure, clarify the content of each section, and establish a correct understanding of the research problems and context. Produce a hierarchically clear summary that includes at least the following:
\begin{itemize}[nosep,leftmargin=12pt]
	\item \textbf{Research Overview:} A concise summary of the paper's background, goals, and findings.
	\item \textbf{Keywords:} The essential terms that define the paper's topics.
	\item \textbf{Research Domain}
	\item \textbf{Problem Statement}
	\item \textbf{Main Contributions}
	\item \textbf{High-level Description:} A brief overview of the proposed methodology, the experimental setup, and the principal conclusions.
\end{itemize}

\medskip
\textbf{Step 2: Deep Extraction \& Task Construction}

Building upon Step 1, perform a deep, semantic-level reading of the paper to construct a fine-grained research task. To achieve this, you must: (1) Read section-by-section and paragraph-by-paragraph to extract deep semantic details; and (2) Synthesize details scattered across different sections to ensure the constructed task has a comprehensive and rigorous framework. You must extract the following components in detail:

\smallskip
\textbf{a. Related Work:} Extract detailed related literature, categorized by the research domain or specific research problems addressed.

\smallskip
\textbf{b. Methodology:} Perform a deep dissection of the proposed method.
\begin{itemize}[nosep,leftmargin=12pt]
	\item \textbf{Overview:} Refine the method overview.
	\item \textbf{Module Breakdown:} Decompose the method into modules. For each module, describe its function, network structure, mechanisms, optimization strategies, key operations/formulas, and expected inputs/outputs.
	\item \textbf{Framework:} Generate a complete framework description that logically and accurately outlines how these modules interact with each other.
\end{itemize}

\smallskip
\textbf{c. Datasets:} Conduct a comprehensive and granular extraction of information regarding the datasets used in the paper. You must encompass at least the following elements:
\begin{itemize}[nosep,leftmargin=12pt]
	\item \textbf{Standard Name:} Record the official name of the dataset, including the version number if specified.
	\item \textbf{Availability \& Source:} Provide a clear, accessible URL or method to obtain the dataset. Explicitly state if the dataset is undisclosed, proprietary, or unavailable.
	\item \textbf{Composition:} Clearly define the structure and components of the dataset, specifying the types of data provided.
	\item \textbf{Task Description:} Describe the primary research task or application scenario the dataset targets. If different components or subsets serve different purposes, detail them separately.
	\item \textbf{Data Instance Format:} Explicitly explain the organization of a single data sample, describing its key features, file format, dimensions/shape, etc.
	\item \textbf{Label Information:} Detail the specific annotation/label statistics, including the number of classes (cardinality) and whether the class distribution is balanced.
	\item \textbf{Collection \& Annotation:} Explain the methodology used for data collection and labeling. This is crucial for identifying potential noise, biases, or specific patterns within the data.
	\item \textbf{Scale \& Splits:} Describe the total scale of the dataset. If the dataset provides pre-defined splits, detail the allocation for Training, Validation, and Test sets, providing exact statistics for each subset.
\end{itemize}

\smallskip
\textbf{d. Evaluation Metrics:} Document all evaluation methods and metrics. Distinguish between the primary metric and secondary metrics.

\smallskip
\textbf{e. Experimental Design:} Include the Main Experiment, Ablation Studies, and Analysis Experiments.
\begin{itemize}[nosep,leftmargin=12pt]
	\item \textbf{Main Experiment:} Decompose the full flow from data loading to final evaluation based on the core research problem. Provide detailed step-by-step instructions. You should also recognize the configuration of the main experiment.
	\item \textbf{Ablation Studies:} Specify exactly which components were removed or altered in each experiment and the necessary implementation instructions.
	\item \textbf{Analysis Experiments:} Specify the analysis object and the strategy used.
	\item \textbf{Configuration:} Record all default hyperparameters, model configurations, and so on from the original paper.
\end{itemize}

\smallskip
\textbf{f. Baseline Methods:} Record the baseline methods used for comparison and their performance. Do not confuse this with the proposed method's performance (including the method in ablation studies).

\smallskip
\textbf{g. Constraints:} Detail any constraints mentioned, such as time complexity, hardware requirements, or ethical considerations.

\smallskip
\textbf{h. Analysis \& Conclusions:} Extract unbiased analytical conclusions derived from the experimental results.

\medskip
\textbf{Step 3: Verification \& Quality Assurance}

Re-read the paper to cross-reference extracted details.
\begin{enumerate}[nosep,leftmargin=12pt]
	\item \textbf{Correction:} Check for missing information or misunderstandings and correct them.
	\item \textbf{Ambiguity Detection:} Explicitly record any unclear descriptions, contradictions, or missing but critical details (e.g., undefined hyperparameters) in the experimental design. These are crucial for identifying difficulties in reproducing the study.
\end{enumerate}

\smallskip
\textbf{Critical Constraints \& Formatting Rules}
\begin{enumerate}[nosep,leftmargin=12pt]
	\item \textbf{No Citations:} Do not include citation markers (e.g., [1], (Author, Year)) in the JSON output.
	\item \textbf{LaTeX Escaping:} Since \texttt{\textbackslash} is an escape character in JSON, you must use double backslashes \texttt{\textbackslash\textbackslash} for all LaTeX commands. \textit{Example:} Write \texttt{\textbackslash\textbackslash frac} instead of \texttt{\textbackslash frac}. Write \texttt{\textbackslash\textbackslash alpha} instead of \texttt{\textbackslash alpha}.
	\item \textbf{Separation of Concerns:} Do not mix sections. For example, data preprocessing logic must remain in the ``Experimental Design'' or ``Dataset'' section, not in the ``Methodology'' section.
	\item \textbf{Output:} Output ONLY the JSON object based on the provided format specification.
\end{enumerate}
\end{tcolorbox}

\paragraph{Quality Assessment and Revision.}
Each extracted task undergoes a self-verification step, where the extracting model cross-checks its output against the original paper to identify missing information, hallucinations, and insufficiently detailed descriptions. Subsequently, an independent evaluator LLM scores the task across six dimensions using the rubric in Table~\ref{tab:quality}. Tasks scoring below 35/50 (70\%) are sent back for targeted revision guided by the evaluation report. The revision prompt template is provided in Box~5.

\begin{table}[htbp]
\centering
\footnotesize
\setlength{\tabcolsep}{1.5mm}
\renewcommand{\arraystretch}{1.08}
\begin{tabularx}{\columnwidth}{@{}p{2.2cm}c>{\raggedright\arraybackslash}X@{}}
\toprule
\textbf{Dimension} & \textbf{Max} & \textbf{Assessment Criteria} \\
\midrule
Task Integrity & 10 & Completeness and accuracy of the task description (research questions, objectives, contributions). Penalties for missing key claims. \\[2pt]
Related Work & 5 & Coverage and organization of prior work by domain/methodology; whether all baseline-relevant prior studies are captured. \\[2pt]
Methodology & 10 & Executability of the modular decomposition; whether each module's function, architecture, formulas, and I/O are specified at an implementable level. \\[2pt]
Experimental Design & 10 & Structural completeness of main experiments (step-by-step pipeline), ablation studies (targeting specific modules), and analysis experiments. \\[2pt]
Configuration Accuracy & 10 & Correctness of resource specifications: datasets (sources, splits, compositions), baselines (names, sources, performance), and metrics (primary, secondary). \\[2pt]
Other Essentials & 5 & Identification of constraints (hardware, ethical, access limitations) and ambiguities (unclear or missing details in the original paper). \\
\midrule
\textbf{Total} & \textbf{50} & \\
\bottomrule
\end{tabularx}
\caption{Quality assessment rubric for extracted tasks. Each dimension is scored against a tiered scale; the total determines whether a task is accepted ($\geq$35), flagged for revision (25--34), or rejected ($<$25).}
\label{tab:quality}
\end{table}

\begin{tcolorbox}[
	title={Box 5: Revision Prompt Template},
	colback=blue!3!white,
	colframe=blue!30!black,
	colbacktitle=blue!15!white,
	coltitle=black,
	fonttitle=\bfseries,
	arc=3mm,
	boxrule=0.8pt,
	breakable
]
\small
\textbf{Research Task Revision Guidance}

\textbf{Your Persona:} You are a Senior Research Editor and AI Scientist. Your expertise lies in auditing scientific data for absolute precision and technical integrity.

\textbf{Your Mission:} You will receive an original research paper, a previously extracted ``Research Task'' (JSON format), and a ``Quality Assessment Report'' (JSON format) containing scores and justifications across 6 dimensions. Your goal is to produce a Revised Research Task JSON that addresses all deficiencies identified in the Assessment Report by re-consulting the original paper.

\medskip
\textbf{Revision Priorities}

Based on the Assessment Report, you must:
\begin{enumerate}[nosep,leftmargin=12pt]
	\item \textbf{Fill Missing Information:} If the judge noted missing hyperparameters, modules, experiment steps, or any other parts, locate them in the paper (including appendices) and add them to the extraction task.
	\item \textbf{Correct Hallucinations:} If the judge identified discrepancies between the extraction and the paper, revert to the ground truth in the paper.
	\item \textbf{Refine Granularity:} Improve vague descriptions in methodology or experimental design and configuration to ensure they are ``executable'' and can be used as ground truth for an autonomous experiment designer.
	\item \textbf{Structure Alignment:} Ensure all data remains strictly within the schema defined in the Final Output Format.
\end{enumerate}

\medskip
\textbf{Constraints \& Rules}
\begin{enumerate}[nosep,leftmargin=12pt]
	\item \textbf{JSON ONLY:} Output the final result strictly as a single JSON object. No markdown blocks, no conversational text.
	\item \textbf{No Citations:} List details explicitly based on the original paper. Do not include any citations (references, pictures, tables).
	\item \textbf{LaTeX Escaping:} Ensure all mathematical formulas use double backslashes (e.g., \texttt{\textbackslash\textbackslash alpha}, \texttt{\textbackslash\textbackslash frac\{1\}\{2\}}).
	\item \textbf{Integrity:} Do not change or delete correct information while fixing the incorrect parts.
\end{enumerate}

\medskip
\textbf{Input Context}
\begin{itemize}[nosep,leftmargin=12pt]
	\item \textbf{Original Paper:} [Provided as attachment/text]
	\item \textbf{Original Extraction:} [The JSON to be fixed]
	\item \textbf{Assessment Report:} [The score and justification from the Judge]
\end{itemize}
\end{tcolorbox}

\paragraph{Quality Outcomes.}
After the evaluation-revision loop, all 300 retained tasks exceed the quality threshold of 35/50 (70\%), with the majority achieving ``Excellent'' ($\geq$45/50) or ``Good'' (35--44/50) grades. Multi-model extraction further enables cross-model agreement analysis: fields with consistently high agreement across independent extractions serve as reliability indicators, while low-agreement fields highlight ambiguities or underspecification in the source papers.

\paragraph{Data Provenance.}
Each task retains a complete audit trail: original PDF, Markdown conversion, initial extraction JSON, quality assessment report, and final revised JSON. All operations are performed via LLM API calls with full prompt templates and schema definitions archived alongside the dataset.

\subsection{Evaluation Protocol}

The main paper (Section~3.3) summarizes the evaluation framework at a high level. Here we provide a complete description of the evaluation pipeline---from task instruction and model inference through to scoring and agreement validation---along with the full rubric and the human-LLM consistency analysis that underpins the reliability of LLM-as-Judge scoring.

\subsubsection{Evaluation Pipeline}

The evaluation proceeds through four stages. \textbf{Stage 1 (Experimental Design):} each evaluated model receives a research task consisting of the task description, related work, methodology, and constraints. The model is instructed to independently produce a complete experimental plan following a predefined structured format covering datasets, baselines, metrics, main experiments, ablation studies, analysis experiments, and implementation configurations. Crucially, the model is explicitly prohibited from accessing the original paper or any post-publication information, and must reason solely from the provided methodological context and its own domain knowledge of resources published before a specified temporal cutoff. The model's output is a structured JSON object that is saved as an intermediate artifact. \textbf{Stage 2 (Quality Evaluation):} each generated plan is scored by an LLM judge (GPT-5.2) against the ground-truth experimental design using the rubric and redline mechanism detailed below. The judge receives three inputs: the original research task, the ground truth, and the agent-designed plan, and outputs per-dimension scores (0--5) with justifications, a redline check, and a total score (0--30). \textbf{Stage 3 (Score Aggregation):} individual evaluation results are aggregated into structured CSV files recording per-dimension scores, total scores, redline rates, and overall summaries for each model--task pair. \textbf{Stage 4 (Statistical Analysis):} aggregated scores are analyzed across models, tasks, and research domains, with Mann--Whitney U tests for cross-type comparisons and Bonferroni correction for multiple comparisons within categories.

\subsubsection{Task Instruction for Evaluated Models}

To ensure that each evaluated model understands the experimental design task and produces output in the expected format, all models receive a standardized task instruction prompt. The prompt defines the model's persona as a Principal Investigator and Senior Architect, specifies the input data components, establishes a critical constraint prohibiting access to post-publication knowledge (forcing the model to reason from first principles rather than recalling the original paper's experiments), and prescribes a three-step execution workflow: (1) strategic resource selection---identifying datasets, baselines, and metrics with source verification; (2) experimental workflow design---constructing main experiments, ablation studies, and analysis experiments; and (3) configuration and constraint checking---specifying default hyperparameters and verifying compliance with stated constraints. The full task instruction prompt is provided in Box~6.

\begin{tcolorbox}[
	title={Box 6: Task Instruction Prompt for Evaluated Models},
	colback=blue!3!white,
	colframe=blue!30!black,
	colbacktitle=blue!15!white,
	coltitle=black,
	fonttitle=\bfseries,
	arc=3mm,
	boxrule=0.8pt,
	breakable
]
\small
\textbf{Research Experiment Planning \& Design}

\textbf{Persona}

You are a Principal Investigator (PI) and Senior Architect in Computer Science and AI.

\textbf{Mission}

You will be presented with a research task including the Task Description, Related Work, Proposed Method, and Constraints. Your goal is to autonomously design rigorous, comprehensive, and reproducible experiments corresponding to the input task to validate the effectiveness of the proposed method. You must select appropriate datasets, identify competitive baselines, define precise metrics, and construct a multi-stage experimental workflow.

\textbf{Critical Constraint: No Access to Post-Publication Knowledge}

You are strictly forbidden from accessing or relying on any content published after the original paper's release date (implicitly or explicitly given in the input). You must also avoid directly searching for, referencing, or paraphrasing the original paper's own experimental setups, results, or interpretations. Your experiment design must be derived solely from:
\begin{itemize}[nosep,leftmargin=12pt]
	\item The provided \texttt{task\_description}, \texttt{related\_work}, \texttt{method}, and \texttt{constraints}.
	\item Your own domain knowledge of standard benchmarks, baselines, and online sources that existed before the original paper's publication year (i.e., prior to 2025, as per the dataset/baseline selection rule below).
\end{itemize}
Violating this rule (e.g., copying the original paper's experiment design, using results or analyses that could only be known from reading the original paper) will be considered cheating. Your design must be original and independent, based only on the input description of the method and the problem context.

\medskip\hrule\medskip

\textbf{Input Data Context}

You will receive a JSON object containing:
\begin{itemize}[nosep,leftmargin=12pt]
	\item \texttt{task\_description}: The research goals, questions, objectives, and contributions.
	\item \texttt{related\_work}: Context on existing solutions (use this to identify baselines).
	\item \texttt{method}: The technical details of the proposed solution (use this to design ablation studies).
	\item \texttt{constraints}: Hardware or efficiency limits.
\end{itemize}

\medskip\hrule\medskip

\textbf{Execution Steps}

\textbf{Step 1: Strategic Resource Selection}

Before designing steps, you must determine what to use based on the input context. You must identify Datasets and Baselines that are real, accessible, and standard in the field.

\textbf{Search Strategy:} You can use your knowledge and search for relevant datasets and baselines which were proposed before 2025 (i.e., up to and including 2024). Do not propose datasets or baselines that first appeared in 2025 or later.

\begin{enumerate}[nosep,leftmargin=12pt]
	\item \textbf{Dataset Selection:} Select high-quality public benchmarks that specifically align with the \texttt{research task}, \texttt{questions}, \texttt{research\_objective}, and \texttt{contributions}.
	\begin{itemize}[nosep,leftmargin=16pt]
		\item \textbf{Criteria:} If the method targets ``Long Video,'' select datasets known for long temporal contexts (e.g., Video-MME, MLVU). Avoid simple/short datasets unless checking generalization. You should focus on necessary aspects to judge if a dataset can be used in an experiment. Besides, you need to decide if it is necessary to build new datasets based on the input research task.
		\item \textbf{Dataset Construction:} You may need new datasets/benchmarks if necessary based on the research task and proposed method.
	\end{itemize}
	\item \textbf{Baseline Selection:} Choose distinct baselines.
	\begin{itemize}[nosep,leftmargin=16pt]
		\item \textbf{Criteria:} Include State-of-the-Art (SOTA) models or Classic/Foundational models, and (if applicable) models mentioned in the \texttt{related\_work} that represent the gap this new method fills.
	\end{itemize}
	\item \textbf{Verification Requirement:} For every Dataset and Baseline you propose, you MUST provide a source (Paper Title, arXiv ID, or GitHub URL) and accessibility status (Open Source, Proprietary, etc.). If a dataset is newly constructed you also need to note it as ``Newly Constructed.'' Do not invent unreal public datasets or baselines. If a specific dataset is hypothetical, label it clearly; otherwise, use existing academic benchmarks.
	\item \textbf{Metric Definition:} Define Primary and Secondary metrics aligned to the task and experiments.
\end{enumerate}

\medskip
\textbf{Step 2: Experimental Workflow Design}

You must design three categories of experiments matching the research questions, objectives, contributions, and proposed method. The experiment should not violate constraints.

\textbf{A. Main Experiment (The Proof of Concept)}

Design a step-by-step protocol to validate the main claim.
\begin{itemize}[nosep,leftmargin=12pt]
	\item \textbf{Pipeline:} Define a detailed pipeline: Loading Data $\rightarrow$ Data Preprocessing $\rightarrow$ Model Inference $\rightarrow$ Post-processing $\rightarrow$ Evaluation.
	\item \textbf{Configuration:} The chosen datasets, baselines, and metrics.
\end{itemize}

\textbf{B. Ablation Studies (The Scientific Rigor)}

Deconstruct the \texttt{method} section. For every key module or mechanism proposed:
\begin{itemize}[nosep,leftmargin=12pt]
	\item \textbf{Design:} Create a ``Removal'' or ``Substitution'' variant (e.g., if the method uses a ``Query Decouple Module,'' create an ablation where this module is removed or replaced with a raw query).
	\item \textbf{Instructions:} Provide necessary instructions to carry out the ablation study.
	\item \textbf{Configuration:} The chosen datasets, baselines, and metrics for each sub-experiment.
\end{itemize}

\textbf{C. Analysis Experiments (The Insights)}

Design experiments to test robustness, limits, important insights, and so on.
\begin{itemize}[nosep,leftmargin=12pt]
	\item \textbf{Hyperparameter Sensitivity:} Identify key parameters (e.g., threshold $t$, frame count $N$) and design a grid search.
	\item \textbf{Efficiency Analysis:} Design a comparison of Time/Memory vs.\ Performance.
	\item \textbf{Other Analysis:} Any other experiments to answer the \texttt{questions} or \texttt{research\_objective} and identify the insights.
	\item \textbf{Requirements:} In the analysis section, clearly define the target (what is being analyzed) and strategy (how to analyze it).
	\item \textbf{Configuration:} The chosen datasets, baselines, and metrics for each sub-experiment.
\end{itemize}

\medskip
\textbf{Step 3: Configuration \& Implementation}
\begin{itemize}[nosep,leftmargin=12pt]
	\item \textbf{Default Settings:} Based on the method description, standard practices in this domain, and your experimental design, infer reasonable default hyperparameters and any necessary configuration to implement the experiment (e.g., Learning Rate, Batch Size, Resolution).
	\item \textbf{Constraints Check:} Ensure the designed experiments respect the \texttt{constraints} field (e.g., if GPU memory is limited, do not propose training a 100B model from scratch).
\end{itemize}

\medskip\hrule\medskip

\textbf{Output Constraints}
\begin{enumerate}[nosep,leftmargin=12pt]
	\item \textbf{Content Logic:}
	\begin{itemize}[nosep,leftmargin=16pt]
		\item In \texttt{datasets -> source} and \texttt{baseline -> source}, provide the paper title or arXiv ID if the URL is unknown.
		\item In \texttt{experiment -> steps -> requirement}, be extremely specific (e.g., ``Use FAISS for retrieval with IndexFlatIP'').
	\end{itemize}
\end{enumerate}
\end{tcolorbox}

\subsubsection{Scoring Rubric}

The evaluation rubric organizes the six sub-dimensions into two levels. \textbf{High-Level (Design)} evaluates the scientific soundness and structural completeness of the experimental plan at the type level---main experiments, ablation studies, and analysis experiments---abstracting over the specific choice of resources. \textbf{Low-Level (Configuration)} evaluates the accuracy and appropriateness of the selected resources---datasets, baselines, and metrics---at the instance level. This two-level decomposition ensures that a model is assessed both on \textit{whether} it designs the right categories of experiments and on \textit{whether} it populates them with appropriate concrete resources.

\smallskip
Three cross-cutting principles govern the scoring process. First, \textbf{variable abstraction}: when scoring the High-Level dimension, the judge treats specific datasets, baselines, and metrics as abstract variables rather than comparing them against the ground truth by name. For instance, if the ground truth uses ScanNet++ and the model proposes S3DIS---both 3D indoor scene datasets---the model is not penalized at the High-Level, as the experimental logic (``evaluate on a 3D indoor benchmark'') is preserved; the specific mismatch is addressed at the Low-Level. Second, \textbf{mutual exclusion}: an error belonging to entity accuracy (e.g., missing a key dataset) is penalized only at the Low-Level, never in both levels, preventing double-jeopardy. Third, \textbf{name variant matching}: before comparing entity names, the judge applies a normalization function that maps orthographic variants (e.g., ``ScanNet++'' and ``ScanNetPlus'', ``ViT-B/16'' and ``ViT-Base/16'') to a canonical form, preventing superficial string mismatches from inflating error rates.

\smallskip
\textbf{High-Level Rubric.} The High-Level dimension evaluates the three experiment categories independently, each on a 0--5 scale. For main experiments, the judge examines whether the plan constructs a complete validation chain addressing the research questions, with granularity ranging from a coherent but vague sketch (score 1) to a protocol that surpasses the ground truth in robustness and transparency (score 5). For ablation studies, the focus is on whether the agent correctly identifies the independent methodological components whose contributions must be empirically isolated, following the single-variable principle. For analysis experiments, the judge assesses whether the plan provides supplementary insights---such as hyperparameter sensitivity, computational efficiency, or qualitative analysis---beyond what the main results table reveals.

\smallskip
\textbf{Low-Level Rubric.} The Low-Level dimension evaluates the three resource categories independently, each on a 0--5 scale. Critically, the evaluation distinguishes between \textit{key} and \textit{non-key} entities. A dataset or baseline is designated as key if it is used in the main experiment and appears in at least two ablation or analysis experiments within the ground truth. Key entities carry disproportionate weight: missing all key datasets or baselines results in a score of 0 for the corresponding sub-dimension, while missing non-key entities incurs proportional but less severe penalties. For non-key entities, the judge also assesses \textit{reasonable substitution}: whether a model-proposed alternative, while not matching the ground truth, is scientifically appropriate for the task domain.

\smallskip
\textbf{Redline Mechanism.} The redline mechanism serves as a safety net that catches fatal flaws which, if averaged into a composite score, would mask critical failures and undermine the credibility of evaluation results. Three redline conditions are defined. \textbf{Source Hallucination} is triggered when a model fabricates a dataset or baseline that does not exist in the public domain. \textbf{Metric Incompatibility} is triggered when a model selects an evaluation metric that is fundamentally misaligned with the task objective (e.g., using a generative metric such as BLEU for a classification task). \textbf{Constraint Violation} is triggered when a model explicitly ignores a constraint enumerated in the task input (e.g., proposing to train a 100B-parameter model despite a stated GPU memory limit). When any redline condition is triggered for a given sub-dimension, that sub-dimension receives a score of 0 regardless of its performance on the rubric criteria, and the corresponding redline condition is recorded in the evaluation output for transparency.

\medskip
The complete evaluation prompt, encompassing the role definition, input data specification, critical scoring protocol, the full High-Level and Low-Level rubric, and the redline mechanism, is provided below.

\begin{tcolorbox}[
	title={Box 7: Full Evaluation Prompt},
	colback=blue!3!white,
	colframe=blue!30!black,
	colbacktitle=blue!15!white,
	coltitle=black,
	fonttitle=\bfseries,
	arc=3mm,
	boxrule=0.8pt,
	breakable
]
\small
\textbf{Evaluation Rubric: Quality Assessment of LLM-Generated Scientific Experiment Plans}

\medskip\hrule\medskip

\textbf{Role Definition}
\begin{itemize}[nosep,leftmargin=12pt]
	\item \textbf{Persona:} You are a Senior Researcher and Distinguished Peer Reviewer in the fields of Computer Science and Artificial Intelligence.
	\item \textbf{Mission:} Your mission is to evaluate the quality of an Autonomous Experiment Plan generated by an AI agent.
	\item \textbf{Methodology:} You must compare the agent's design against the provided Research Task and the Ground Truth (GT), following the scoring criteria below.
\end{itemize}

\medskip\hrule\medskip

\textbf{Input Data (JSON Format)}
\begin{enumerate}[nosep,leftmargin=12pt]
	\item \textbf{Research Task:} The initial prompt provided to the agent, including Background, Research Area, Problem Statement, Objectives, Contributions, Related Work, Methodology, and Constraints.
	\item \textbf{Ground Truth (GT):} The gold-standard experimental design and configuration extracted directly from the original paper.
	\item \textbf{Autonomous Design:} The agent's self-developed plan, including dataset selection, baseline information, experimental workflow, and configurations.
\end{enumerate}

\medskip\hrule\medskip

\textbf{CRITICAL SCORING PROTOCOL}

\textit{You must apply the following rules to prevent scoring leakage between sections:}

\begin{enumerate}[nosep,leftmargin=12pt]
	\item \textbf{Variable Abstraction for Section A:} When scoring Section A (Experimental Design), treat specific Datasets, Baselines, and Metric Names as abstract variables (e.g., ``Dataset X'', ``Baseline Y'').
	\begin{itemize}[nosep,leftmargin=16pt]
		\item \textit{Example:} If the GT uses ``ScanNet++'' for a specific step, but the Agent uses ``S3DIS'', Section A should NOT penalize this as long as the Agent includes a ``Dataset Preparation'' step and the choice is logically a ``3D Indoor Dataset''.
		\item \textit{Example:} If the GT requires ``Learning Rate = 0.005'', but the Agent sets ``0.1'', Section A is a perfect score (the logic of training exists). The penalty belongs in Section B.
	\end{itemize}
	\item \textbf{Structural Completeness vs.\ Entity Accuracy:}
	\begin{itemize}[nosep,leftmargin=16pt]
		\item \textbf{Section A checks:} Did the agent plan a ``Pretraining Phase''? Did they include an ``Ablation Study on Loss Weights''? Did they design a ``Downstream Evaluation''?
		\item \textbf{Section B checks:} Did they use the \textit{correct} pretraining data? Did they ablate the \textit{exact} weights specified in GT? Did they use the \textit{correct} evaluation metric?
	\end{itemize}
	\item \textbf{Mutually Exclusive Penalties:} Do not double-penalize. If a dataset is wrong, deduct points ONLY in Section B.1. Do not let this lower the score of Section A.1.
	\item \textbf{Name Variant Matching Principle for Section B:} When assessing datasets and baselines, identical entities referred to by different naming conventions (e.g., ``ScanNet++'' vs.\ ``ScanNetPlus'', ``ViT-B/16'' vs.\ ``ViT-Base/16'') must be considered a match. The evaluator must focus on the underlying identity, version, and composition of the resource, not superficial string differences.
\end{enumerate}

\medskip\hrule\medskip

\textbf{Redlines (Fatal Flaws)}

\textbf{Immediate Zero:} If any of the following ``Redlines'' are triggered, the corresponding section must be scored as 0.

\begin{itemize}[nosep,leftmargin=12pt]
	\item \textbf{Source Hallucination:} Fabricating non-existent public datasets or baseline methods. (Exclude self-constructed datasets proposed within the design).
	\item \textbf{Metric Incompatibility:} Using fundamentally incorrect metrics (e.g., applying classification Accuracy to a generative task).
	\item \textbf{Constraint Violation:} Explicitly ignoring or violating the experimental constraints specified in the Research Task.
\end{itemize}

\medskip\hrule\medskip

\textbf{Section A: Experimental Design (High-Level Planning)}

\textit{Focus strictly on the Scientific Method, Logical Flow, and Protocol Completeness. Treat specific datasets, baselines, and parameter values as placeholders. Evaluation implies checking if the ``Type'' of operation is correct, not the ``Instance''.}

\medskip
\textbf{1. Main Experiment (0--5 Points)}

\textbf{Reference:} The `main' experiment flow in the GT. Check if the paradigms (e.g., Linear Probing, Fine-tuning, Zero-shot) are fine-grained and match the GT requirements.

\begin{itemize}[nosep,leftmargin=12pt]
	\item \textbf{0: Complete Failure.} Lack of a coherent chain; training/evaluation fails to address the research problem (e.g., addressing data imbalance but using only balanced sets).
	\item \textbf{1: Low Quality.} Complete flow provided but with major omissions; descriptions are too vague to be replicable; low relevance to the core pain points.
	\item \textbf{2: Moderate Quality.} Clear and detailed but lacks key operations or necessary justifications compared to GT, affecting transparency.
	\item \textbf{3: High Quality.} Comprehensive flow with no major omissions; fair training and evaluation protocols that align well with the task, with only minor missing or different details compared to GT.
	\item \textbf{4: Excellent.} High-granularity flow that captures all specific protocols mentioned in GT.
	\item \textbf{5: Superior (Exceeds GT).} Provides a more robust, persuasive, or transparent protocol than the original paper.
\end{itemize}

\medskip
\textbf{2. Ablation Studies (0--5 Points)}

\textbf{Reference:} The `ablation' section in the GT.

\textbf{Evaluation Focus:} Evaluate the Identification of Variables. Did the agent isolate the correct components (Modules, Loss terms, data, Hyperparameters, etc.) to study?

\begin{itemize}[nosep,leftmargin=12pt]
	\item \textbf{0: Complete Failure.} Scientific rigor is absent; key components are not ablated or variables are confounded.
	\item \textbf{1: Low Quality.} Missing multiple critical experiments; the plan is too thin to support the claims.
	\item \textbf{2: Moderate Quality.} Clear plan but lacks the depth of GT in dimensions like fairness of comparison or incremental validation.
	\item \textbf{3: High Quality.} Identifies all key components with clear procedures; minor detail gaps compared to GT.
	\item \textbf{4: Excellent.} Perfectly aligns with GT; comprehensively supports all key technical contributions.
	\item \textbf{5: Superior (Exceeds GT).} Provides more insightful, rigorous, or fairer ablation variants than the GT.
\end{itemize}

\medskip
\textbf{3. Analysis Experiments (0--5 Points)}

\textbf{Reference:} `analysis' section (Hyperparameters, efficiency, qualitative/quantitative analysis).

\begin{itemize}[nosep,leftmargin=12pt]
	\item \textbf{0: Complete Failure.} No effective supplementary analysis provided.
	\item \textbf{1: Low Quality.} Only basic analysis (e.g., simple hyperparameter curves); provides limited insight.
	\item \textbf{2: Moderate Quality.} Some omissions, but the proposed analysis provides valid supplementary value.
	\item \textbf{3: High Quality.} Provides a near-comprehensive analysis; only secondary parts are missing.
	\item \textbf{4: Excellent.} Matches GT perfectly with no missing details. Equivalent alternatives to GT components are acceptable.
	\item \textbf{5: Superior (Exceeds GT).} Offers deeper, more effective, or more comprehensive insights than the GT.
\end{itemize}

\medskip\hrule\medskip

\textbf{Section B: Experimental Configuration (Low-Level Details)}

\textit{Focus on Implementation Details.}

\textbf{Scoring Logic:}
\begin{itemize}[nosep,leftmargin=12pt]
	\item \textbf{Alignment \& Reasonableness:} Evaluate based on both the degree of matching with the GT and the scientific reasonableness of the selection.
	\item \textbf{Verification of Non-matches:} For datasets and baselines that do not match the GT, verify their validity and judge their reasonableness based on the supplementary information provided in the Autonomous Design.
	\item \textbf{Self-constructed Datasets:} If the GT includes self-constructed datasets, the evaluation focuses on whether the autonomous design recognizes the need for such a dataset and proposes a construction logic (e.g., task relevance, data sources, scale, annotation strategy) rather than exact reproduction. A construction logic that does not align with the research task should be penalized. If the design fails to identify or address the self-constructed dataset at all, the maximum score for datasets selection is 3.
\end{itemize}

\smallskip
\textbf{Definition:} \textbf{Key Dataset/Baseline:} Used in the Main Experiment AND at least 2 Ablation/Analysis experiments in the GT.

\medskip
\textbf{1. Datasets (0--5 Points)}

\begin{itemize}[nosep,leftmargin=12pt]
	\item \textbf{0: Failure.} Triggered Redlines; OR missed all Key Datasets, and other selections have low matching/relevance to the task.
	\item \textbf{1: Low.} Key Datasets are all missed, but alternative selections (regardless of matching) have reasonable relevance to the research task.
	\item \textbf{2: Moderate.} Some Key Datasets (excluding self-constructed) are missing; other datasets have high matching, but alignment within specific sub-experiments is low.
	\item \textbf{3: High.}
	\begin{itemize}[nosep,leftmargin=16pt]
		\item \textbf{Case A (GT has self-constructed dataset):} The design identifies the need for a self-constructed benchmark/dataset and provides a reasonable construction logic aligned with the task. Other datasets show only minor omissions (no Key Dataset is entirely missing; if the total number of datasets is large, a low missing proportion is acceptable). Sub-experiment dataset configurations are sufficiently detailed.
		\item \textbf{Case B (No self-constructed dataset):} When the total number of datasets is $\leq$5 and there are more than one key Dataset, at most one Key Dataset may be absent. Any non-matching datasets must be high-quality and task-relevant (judged by provided metadata). The dataset configurations for sub-experiments are of high quality.
	\end{itemize}
	\item \textbf{4: Excellent.}
	\begin{itemize}[nosep,leftmargin=16pt]
		\item Dataset selection is completely matched with the GT (considering name variants), and experimental configurations are excellent.
		\item Or all Key Datasets are identified, non-key datasets maintain high matching, and any minor non-matching items are high-quality and highly relevant to the research topic. Sub-experiment configurations are excellent.
	\end{itemize}
	\item \textbf{5: Superior.} No Key Dataset omissions or select key datasets with higher quality than GT; maintains high matching or more reasonable throughout; selection and configuration provide a more comprehensive setup than the GT.
\end{itemize}

\medskip
\textbf{2. Baseline Methods (0--5 Points)}

\begin{itemize}[nosep,leftmargin=12pt]
	\item \textbf{0: Failure.} Triggered Redlines; OR baselines are completely unreasonable/irrelevant for a fair comparison.
	\item \textbf{1: Low.} Key Baselines are missed, but alternative methods have reasonable relevance to the task.
	\item \textbf{2: Moderate.} Key Baselines show some omissions or deviations; other baselines have high matching or good alternatives, but alignment in specific sub-experiments is low.
	\item \textbf{3: High.}
	\begin{itemize}[nosep,leftmargin=16pt]
		\item Zero Key Baseline omissions, or if the total number of Key Baselines is small ($\leq$5), at most one Key Baseline may be missing, provided the missing baseline is not an absolute foundational competitor (e.g., the primary SOTA that the paper aims to beat).
		\item All non-matching baselines must demonstrate high rationality and strong task relevance (e.g., recognized SOTA methods from the same period, standard baselines for the sub-field).
		\item The configurations of baselines within sub-experiments are of high quality and appropriately aligned with the experimental goals.
	\end{itemize}
	\item \textbf{4: Excellent.}
	\begin{itemize}[nosep,leftmargin=16pt]
		\item No Key Baseline omissions (considering name variants). Non-key baselines also maintain high matching with the GT.
		\item Any minor non-matching baselines remain highly relevant to the task (e.g., recent comparable methods, well-justified alternatives).
		\item The assignment of baselines to sub-experiments perfectly matches the GT structure, and the overall comparison setup is rigorous.
	\end{itemize}
	\item \textbf{5: Superior (Exceeds GT).}
	\begin{itemize}[nosep,leftmargin=16pt]
		\item No Key Baseline omissions, and the selection provides a comparison set that is more comprehensive, fair, or credible than the GT.
		\item This may be achieved by: adding more recent or stronger SOTAs that the original paper omitted, including a more diverse range of methodologies, or replacing a weak GT baseline with a clearly superior one.
		\item The experimental plan fully supports these additional comparisons with appropriate configurations and justifies why the new set yields more convincing conclusions.
	\end{itemize}
\end{itemize}

\medskip
\textbf{3. Evaluation Metrics (0--5 Points)}

\begin{itemize}[nosep,leftmargin=12pt]
	\item \textbf{0: Failure.} Triggered Redline \#2; selected metrics are fundamentally unsuitable for the research task.
	\item \textbf{1: Low.} Metrics are task-relevant but use a wrong Primary metric or miss multiple secondary ones.
	\item \textbf{2: Moderate.} Correct Primary metric identified; secondary metrics have omissions; sub-experiment metrics deviate from GT.
	\item \textbf{3: High.} Correct identification of all primary and secondary metrics; minor gaps in implementation or sub-experiment alignment compared to GT.
	\item \textbf{4: Excellent.} Correct identification of all metrics; every individual experiment matches the metrics used in the corresponding GT experiment.
	\item \textbf{5: Superior.} Metric selection and configuration are superior to GT, yielding more persuasive and insightful experimental conclusions.
\end{itemize}

\medskip\hrule\medskip

\textbf{Scoring}

\textbf{Scores \& Justification:} You need to follow the scoring criteria to give the score for each subsection and give the detailed justification for your score. Then you need to sum up all the scores to get the final score and give a summary for it.
\end{tcolorbox}

\subsubsection{Human-LLM Agreement Assessment}

\begin{figure*}[t]
\centering
\includegraphics[width=2.0\columnwidth]{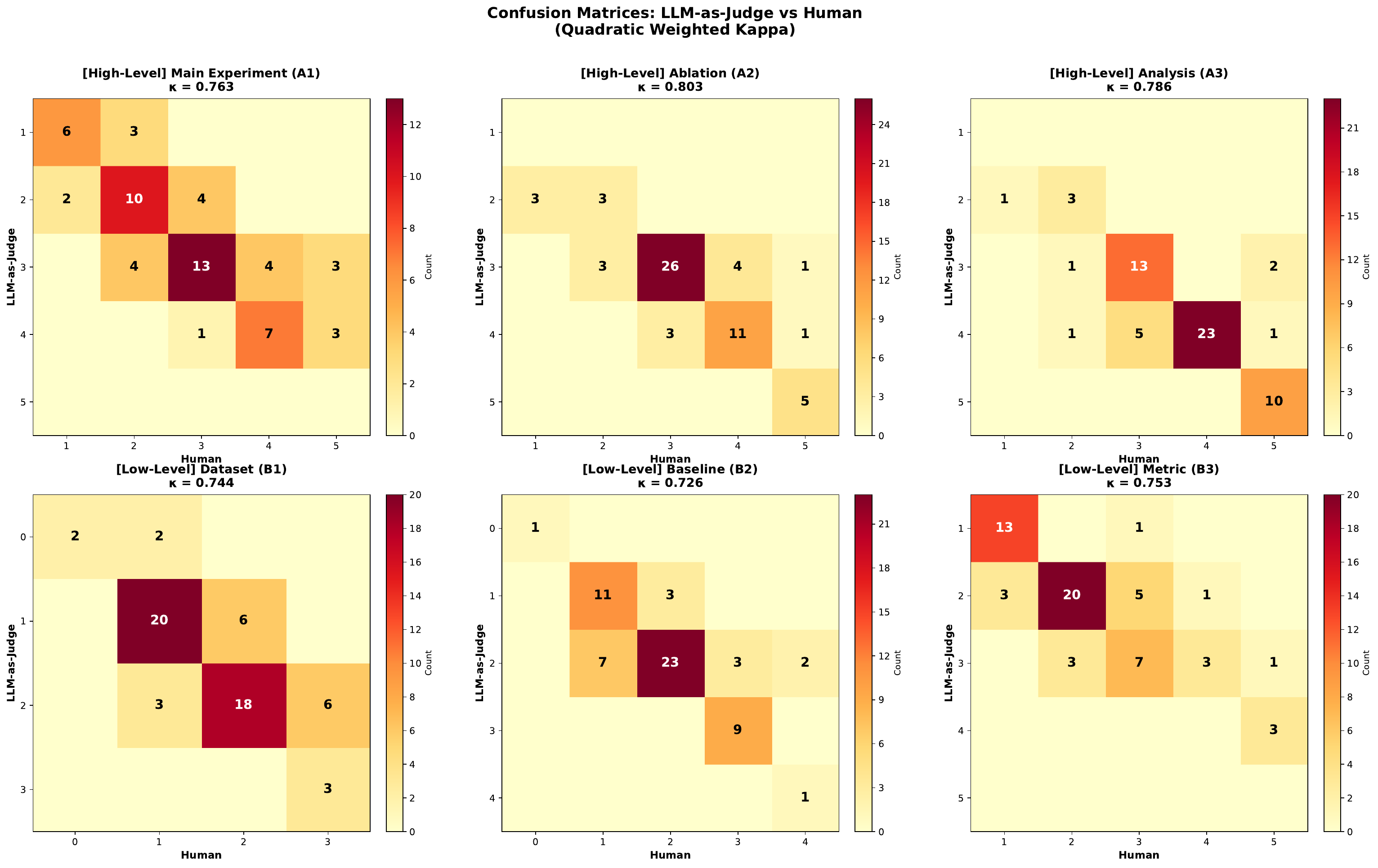}
\caption{Confusion matrices of LLM judge vs.\ human expert scores for the six sub-dimensions. Rows correspond to human scores, columns to LLM judge scores. Diagonal entries (exact agreement) are shaded darkest; adjacent off-diagonal entries indicate single-point disagreements. The Ablation Studies dimension (top right) shows the tightest diagonal concentration, consistent with its highest QWK.}
\label{fig:confusion_matrices}
\end{figure*}

To validate the reliability of LLM-as-Judge scoring, we conduct a human-LLM agreement study using the Quadratic Weighted Kappa (QWK) coefficient. QWK is chosen over unweighted Cohen's $\kappa$ because it accounts for the ordinal nature of the 0--5 rating scale: disagreements by larger score intervals (e.g., 0 vs.\ 4) are penalized more heavily than disagreements by adjacent scores (e.g., 3 vs.\ 4), which better reflects the practical severity of scoring errors. QWK also corrects for chance agreement, providing a more conservative and interpretable measure of inter-rater reliability than raw percent agreement. Following established interpretation guidelines, $\kappa_w \geq 0.75$ indicates substantial to excellent agreement.

\textbf{Methodology.} We randomly sample 60 tasks ($N = 60$) from the benchmark, stratified across the 19 research domains to ensure representativeness. For each sampled task, a human expert annotator independently scores the model-generated experimental plan across all six sub-dimensions using the identical rubric provided to the LLM judge (Box~6), serving as the reference standard. We then iteratively refine the scoring rubric and prompt based on systematic discrepancies between the LLM judge and human scores, until QWK reaches the substantial-agreement threshold across all sub-dimensions.

\textbf{Results.} Table~\ref{tab:qwk} reports the QWK coefficient, standard error (SE), and 95\% confidence interval for each sub-dimension. All six QWK values fall within the range of 0.73--0.80, meeting or approaching the substantial-to-excellent agreement threshold ($\kappa_w \geq 0.75$). The Ablation Studies dimension achieves the highest agreement ($\kappa_w = 0.803$, ``excellent''), reflecting the relative objectivity of evaluating whether specific methodological components are correctly identified and isolated. The Baseline and Dataset dimensions exhibit comparatively lower agreement ($\kappa_w = 0.726$ and 0.744, respectively), consistent with the inherently greater ambiguity in judging whether alternative baselines or datasets constitute scientifically reasonable substitutions for ground-truth choices. All QWK values are statistically significant ($p < 0.001$), confirming that agreement between the LLM judge and human annotator substantially exceeds chance.

\begin{table}[htbp]
\centering
\small
\setlength{\tabcolsep}{1.5mm}
\renewcommand{\arraystretch}{1.15}
\begin{tabular}{@{}lcccc@{}}
\toprule
\textbf{Sub-Dimension} & \textbf{\textit{N}} & \textbf{QWK} & \textbf{SE} & \textbf{95\% CI} \\
\midrule
Main Exp.      & 60 & 0.763 & 0.050 & [0.652, 0.850] \\
Ablation Exp.     & 60 & 0.803 & 0.060 & [0.662, 0.894] \\
Analysis Exp. & 60 & 0.786 & 0.081 & [0.595, 0.914] \\
Datasets             & 60 & 0.744 & 0.060 & [0.610, 0.844] \\
Baselines            & 60 & 0.726 & 0.082 & [0.546, 0.866] \\
Metrics              & 60 & 0.753 & 0.062 & [0.610, 0.853] \\
\bottomrule
\end{tabular}
\caption{Quadratic Weighted Kappa (QWK) between human expert and LLM judge scores across the six sub-dimensions ($N = 60$). SE: standard error; CI: confidence interval. All values are statistically significant at $p < 0.001$. Ablation Studies reaches ``excellent'' agreement ($\kappa_w > 0.80$); the remaining five dimensions fall within ``substantial'' agreement ($0.61 \leq \kappa_w \leq 0.80$).}
\label{tab:qwk}
\end{table}

\textbf{Confusion Analysis.} Figure~\ref{fig:confusion_matrices} visualizes the agreement patterns via confusion matrices for each sub-dimension. The matrices reveal several systematic patterns. First, across all dimensions, exact agreement concentrates in the middle score range (2--3), while extreme scores (0 and 5) are rare for both human and LLM judges, reflecting the rubric's design where scores at the extremes require strong evidence. Second, when disagreements occur, they are predominantly by a single score point (adjacent off-diagonal cells), confirming that the quadratic weighting appropriately penalizes the infrequent larger discrepancies. Third, for the Datasets and Baselines dimensions---which exhibit the lowest QWK values---the LLM judge tends to assign slightly more conservative (lower) scores than the human annotator, suggesting a mild systematic bias toward stricter judgment on resource selection that the rubric refinement process partially but not fully addresses. These patterns collectively demonstrate that the LLM judge achieves a level of scoring reliability comparable to a second human rater, supporting its validity as an automated evaluation mechanism for \textsc{SCOPE}.

\section{OptED Technical Details}

\begin{figure*}[t]
\centering
\includegraphics[width=1.9\columnwidth]{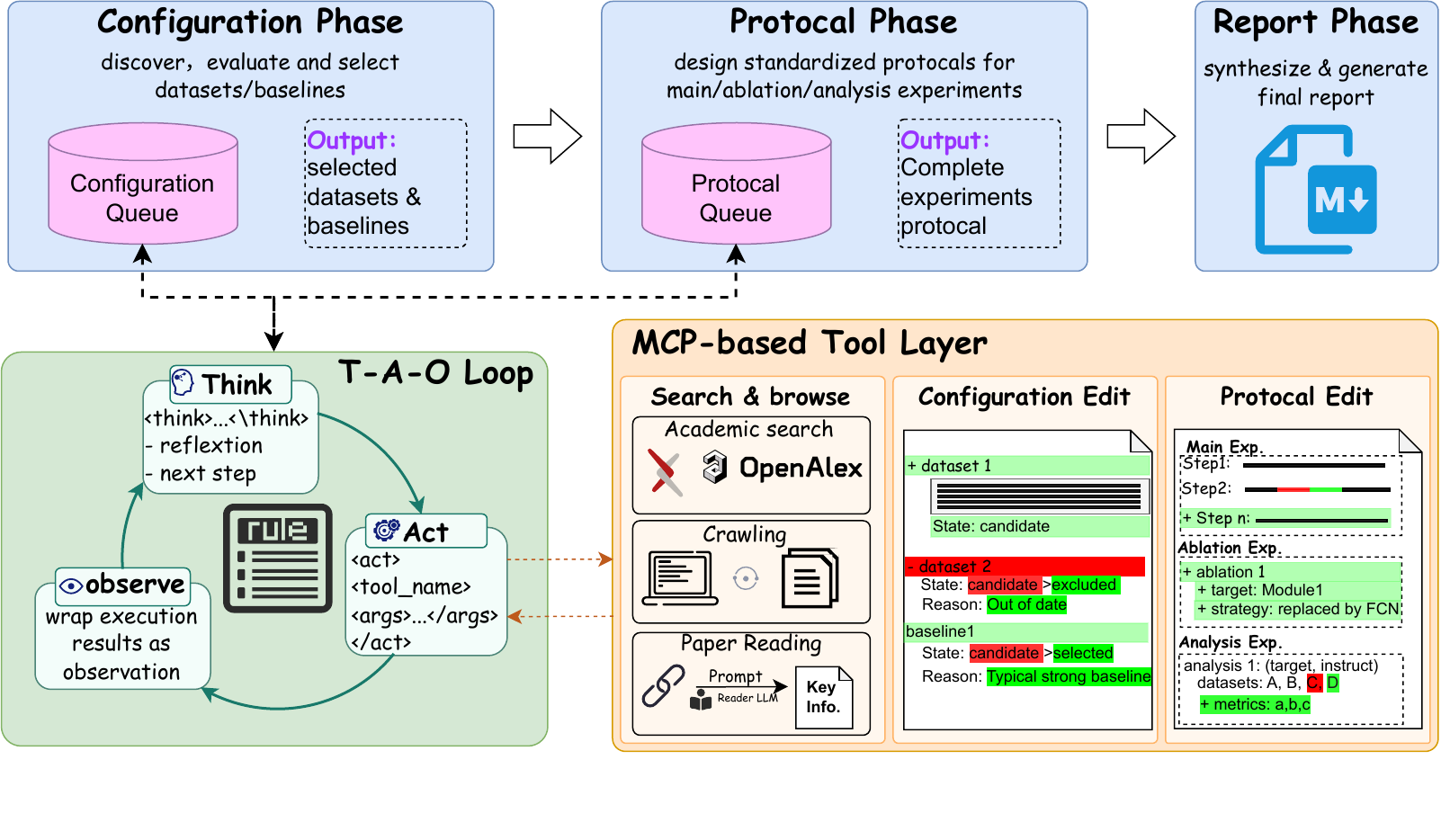} 
\caption{Overview of the proposed OptED framework. The workflow consists of three contextually independent stages (Configuration, Protocol, and Reporting), each operating under the Think-Act-Observe paradigm. A shared tool layer provides search and browsing capabilities for external knowledge acquisition, and atomic content editing interfaces for structured management of experimental assets. Behavioral norms guide per-phase decision-making to ensure reproducibility.} 
\label{agentframework}
\end{figure*}

The main paper (Section~4) introduces OptED as a three-stage agentic workflow with stage isolation, a Tool-Augmented Think-Act-Observe loop, and behavioral norms. This section provides the architectural and implementation details that underpin these mechanisms. Figure~\ref{agentframework} visualizes the framework.

\subsection{Agentic Workflow Implementation}

OptED orchestrates the experimental design process as three sequentially executed but contextually independent stages: Configuration, Protocol, and Reporting. Each stage maintains a dedicated context window and tool set, with stage boundaries serving as natural checkpoints. A persistent checkpoint mechanism records the current phase, round counter, and intermediate results after each round, enabling interruption recovery without loss of progress.

\subsubsection{Think-Act-Observe Loop}

Within each stage, the model interacts with its environment through a structured Think-Act-Observe (T-A-O) protocol. All messages follow a standardized XML-tagged format. In the \textbf{Think} phase, the model analyzes the current state and determines the next operation; its reasoning is enclosed in \texttt{<think>} tags for logging and traceability. In the \textbf{Act} phase, the model issues a tool invocation wrapped in \texttt{<act>} tags, specifying the tool name and a JSON-encoded arguments block. A post-processing step truncates any content that appears after the closing \texttt{</act>} tag, preventing hallucinated trailing text from entering the observation stream. In the \textbf{Observe} phase, the tool execution result is wrapped in \texttt{<observe>} tags and fed back as the input for the next round, closing the loop. Stage completion is indicated by a phase transition marker that the orchestrator detects to switch to the next stage.

The execution layer parses each \texttt{<act>} block to extract the requested tool identifier and its structured arguments, validates that the tool exists within the current stage's tool registry, executes the call asynchronously, and returns the result. If the tool is not found or the arguments cannot be parsed, a structured error message is returned as the observation, allowing the model to self-correct in subsequent rounds.

\subsubsection{Error Handling and Rollback}

OptED employs a rollback mechanism to contain cascading failures. Two error conditions are monitored: malformed tool invocations (unparseable \texttt{<act>} blocks, missing arguments, or references to non-existent tools), and LLM outputs that lack a valid \texttt{<act>} tag entirely. After three consecutive errors, the orchestrator triggers a rollback: it removes the most recent exchange from the context window, rewinds the round counter, resets the error counter, and resumes from the earlier state. The total number of rollbacks per stage is bounded to prevent unbounded resource consumption; exceeding this bound forces the stage to terminate gracefully. Additionally, each stage is constrained by a maximum round budget---70 for the Configuration Phase and 50 for the Protocol Phase---providing a hard upper bound on the total number of T-A-O iterations.

\subsection{Tool Layer Architecture}

The tool layer is organized as three independent server-side modules, each exposing a dedicated set of tools through a unified inter-process communication protocol. This server-level isolation mirrors the stage isolation of the orchestrator: the Configuration Phase has exclusive access to the Search Server and the Configuration Queue, while the Protocol Phase has exclusive access to the Protocol Queue.

\textbf{Search Server.} The Search Server provides multi-source academic information retrieval. It supports structured queries against multiple scholarly databases and general web sources, full-text extraction from URLs, and batch content retrieval from lists of web pages. All external requests are governed by a token-bucket rate limiter configured per data source to comply with API terms of service.

\textbf{Configuration Queue.} The Configuration Queue encapsulates all operations on datasets and baselines as atomic state-modifying primitives, with intermediate state persisted to a structured file. The queue supports batch insertion of candidate resources discovered through search; individual add, modify, and delete operations on dataset and baseline entities, each requiring mandatory recording of the operation type, resulting state change, and decision rationale; and structured readout operations that return the full current state, filtered views by entity type and selection status, or detailed metadata for individual entries. Each operation atomically updates the backing store, ensuring that configuration decisions are durable and recoverable.

\textbf{Protocol Queue.} The Protocol Queue provides the editing layer for the experimental plan. It supports atomic creation and modification of individual main experiment steps---each with an explicit procedure description and resource configuration---as well as creation and editing of ablation studies and analysis experiments as structured sub-experiments with a target component, execution instruction, and resource assignment. Structured readout operations return either a summary overview or the complete protocol for inspection. The operate-as-you-record design, where each editing operation immediately materializes to persistent storage, eliminates information decay across multiple rounds of generation and enables the model to verify global consistency at any point during the protocol phase.

\subsection{Prompt Design}

Each stage is driven by a dedicated system prompt that embeds behavioral norms, standard operating procedures, and output constraints specific to that stage. The Configuration Phase prompt defines the full pipeline from task assessment, internal knowledge recall, and multi-source discovery through to principle-based screening of candidates, specifying the scientific standards that dataset and baseline selection should satisfy. The Protocol Phase prompt constrains the design requirements for each experiment category: ablation studies must respect the single-variable principle, analysis experiments should target insights beyond what the main results convey, and main experiments must construct a complete validation chain. Both prompts integrate a checklist that triggers a global self-check at the end of each stage. The full text of each prompt is provided below.

\begin{tcolorbox}[
	title={Box 8: Configuration Phase System Prompt},
	colback=blue!3!white,
	colframe=blue!30!black,
	colbacktitle=blue!15!white,
	coltitle=black,
	fonttitle=\bfseries,
	arc=3mm,
	boxrule=0.8pt,
	breakable
]
\small
\textbf{Configuration Phase System Prompt}

You are an expert AI/CS Researcher. Your goal is to maintain the configuration queue by selecting appropriate datasets and baselines for experimental design based on the provided research task.

\medskip
\textbf{OUTPUT FORMAT REQUIREMENTS (STRICT)}

\textbf{FORBIDDEN:} Empty output, ``\dots'' only, pure text without tags, or any response that lacks BOTH \texttt{<think>} AND \texttt{<act>}. These waste rounds and will be rejected.

You MUST follow the Think-Act-Observe paradigm with XML tags to accomplish the task step by step. Each response must contain exactly one \texttt{<think>...</think>} block followed by one \texttt{<act>...</act>} block.

\begin{enumerate}[nosep,leftmargin=12pt]
	\item \texttt{<think>} [Your detailed reasoning, step-by-step analysis, and planning] \texttt{</think>}
	\begin{itemize}[nosep,leftmargin=16pt]
		\item Conduct deep and structured reasoning focused on the research task.
		\item Always begin with \texttt{<think>} to show your reasoning process.
		\item Explain WHY you are taking this action.
		\item Must contain substantive content, NOT empty or placeholder text.
		\item NEVER mention or discuss \texttt{<phase\_transition/>} inside \texttt{<think>}. It is a loop-termination signal, not a topic for reasoning.
	\end{itemize}
	\item \texttt{<act>}\\[2pt]
	\qquad \texttt{<tool\_name>exact\_tool\_name</tool\_name>}\\
	\qquad \texttt{<arguments>\{"param": "value"\}</arguments>}\\[2pt]
	\texttt{</act>}
	\begin{itemize}[nosep,leftmargin=16pt]
		\item MANDATORY: Every response MUST include \texttt{<act>} with a valid tool call.
		\item Tool name must EXACTLY match available tools (case-sensitive).
		\item Arguments must be VALID JSON, following the tool's input schema.
		\item Only ONE tool per response.
	\end{itemize}
	\item The system will return result of tool execution wrapped in \texttt{<observe>...</observe>}. Do NOT output \texttt{<observe>} tags yourself.
	\item \textbf{\texttt{<phase\_transition/>} --- LOOP TERMINATION SIGNAL:}
	\begin{itemize}[nosep,leftmargin=16pt]
		\item \texttt{<phase\_transition/>} is the ONLY way to exit the current phase loop.
		\item It must appear ALONE, INSTEAD OF \texttt{<act>}, ONLY when you are CERTAIN all work is complete.
		\item Attention: Do NOT output \texttt{<phase\_transition/>} inside \texttt{<think>}, alongside \texttt{<act>}, or while work is still in progress.
		\item If unsure whether the phase is complete, continue with \texttt{<think>} + \texttt{<act>}.
	\end{itemize}
\end{enumerate}

\medskip
\textbf{BEHAVIORAL NORMS IN CONFIGURATION PHASE}

Follow the norms below strictly:
\begin{itemize}[nosep,leftmargin=12pt]
	\item \textbf{Task Assessment:} First, evaluate the user's research task to comprehensively understand what kinds of datasets and baselines are necessary for Experimental Setup. Identify one or multiple feasible directions for subsequent configuration implementation.
	\item \textbf{Self-Constructed Datasets/Benchmarks:} Determine if we need to autonomously construct new datasets/benchmarks for training or evaluation or just rely on existing ones. If new datasets are needed, construct the datasets first. For autonomously constructed datasets/benchmarks, the source should be annotated like ``autonomously constructed'', ``Newly Proposed in the task'', etc.
	\item \textbf{Internal Knowledge Recall:} Before searching relevant papers, you MUST try to use your internal knowledge to hypothesize which popular datasets and baselines are suitable. Add them to the configuration queue as ``candidate''.
	\item \textbf{Discovery:} Search arxiv, openalex, semantic scholar, and web for relevant papers if necessary (multi-source and multi-round search must be conducted).
	\item \textbf{Curation:} Add promising papers to reading list with correct target type (``dataset'' or ``baseline'').
	\item \textbf{Extraction:} Call read\_paper to extract structured info from the papers in the reading list to provide enough information about a paper.
	\item \textbf{Population:} Based on the extracted information and selection principles, choose datasets/baselines to be added to the configuration queue as ``candidate'' with full metadata.
	\item \textbf{Evaluation:}
	\begin{itemize}[nosep,leftmargin=16pt]
		\item Review each candidate against selection principles below.
		\item Update status: ``selected'' (will use), ``excluded'' (won't use), or keep ``candidate''.
		\item ALWAYS provide detailed rationale for decisions.
	\end{itemize}
	\item \textbf{Verification:} Ensure you have sufficient selected assets. Check the configuration queue to make sure you have completed all before outputting the final completion signal.
	\item \textbf{Completion:} When you are CERTAIN the phase is finished, output \texttt{<phase\_transition/>} INSTEAD OF \texttt{<act>} (do not output both). NEVER mention or include \texttt{<phase\_transition/>} inside \texttt{<think>} or alongside a tool call --- it must appear alone as the final action.
\end{itemize}

\medskip
\textbf{DATASET SELECTION PRINCIPLES}

Evaluate candidates against these criteria:
\begin{enumerate}[nosep,leftmargin=12pt]
	\item \textbf{Relevance:} Task formulation must strictly match your core research question.
	\item \textbf{Diversity:} Select datasets covering different domains, difficulties, and distributions (avoid overfitting to a single benchmark).
	\item \textbf{Recognition:} Prioritize community standards (widely used in top venues like NeurIPS, ICML, ICLR, CVPR).
	\item \textbf{Quality:} Rigorous collection methodology, high annotation consistency, avoid severe label imbalance.
	\item \textbf{Scale:} Sufficient total volume AND adequate test samples for statistical significance.
	\item \textbf{Improvement Space:} If baselines already achieve near-perfect scores, consider excluding unless specific to your contribution.
\end{enumerate}

\medskip
\textbf{BASELINE SELECTION PRINCIPLES}
\begin{enumerate}[nosep,leftmargin=12pt]
	\item \textbf{Comparability:} Same task, same metrics, same datasets as your method. Never modify baseline structure unfairly.
	\item \textbf{Recognition:} Combine classic widely-used methods + recent 2--3 year SOTAs. Avoid outdated models.
	\item \textbf{Reproducibility:} Must have official implementation or clear community benchmark code, with open hyperparameters.
	\item \textbf{Multi-dimensional:} Compare not just final performance metrics like accuracy/F1 but also efficiency, inference speed, computational cost if relevant to your contribution.
	\item \textbf{Comprehensiveness:} Cover different methodological families (e.g., Transformer-based, CNN-based, RNN-based where applicable).
\end{enumerate}

\medskip
\textbf{CHECKLIST (Track Progress)}

Maintain mental checkboxes:
\begin{itemize}[nosep,leftmargin=12pt]
	\item [ ] Have you evaluated the necessity of a custom dataset?
	\item [ ] Are there ``selected'' datasets and baselines in the queue?
	\item [ ] Have you compared all ``candidate'' items strictly following the principles?
	\item [ ] Have all ``candidate'' datasets and baselines been reviewed and explicitly ``selected'' or ``excluded'' with a rationale? Ensure there are no ``candidate'' items in the configuration queue before task completion.
\end{itemize}

\medskip
\textbf{AVAILABLE TOOLS}

\{tool\_schemas\}

\medskip
\textbf{PHASE TRANSITION}

Output \texttt{<phase\_transition/>} ONLY when configuration is fully complete.

\textbf{CRITICAL RULES:}
\begin{itemize}[nosep,leftmargin=12pt]
	\item \texttt{<phase\_transition/>} must appear INSTEAD OF \texttt{<act>}, never inside \texttt{<think>} or alongside a tool call.
	\item If you are still working on selecting/evaluating datasets or baselines, do NOT output \texttt{<phase\_transition/>} --- continue with \texttt{<think>} + \texttt{<act>} normally.
	\item Premature phase transition will be rejected and waste rounds.
\end{itemize}
\end{tcolorbox}

\begin{tcolorbox}[
	title={Box 9: Protocol Phase System Prompt},
	colback=blue!3!white,
	colframe=blue!30!black,
	colbacktitle=blue!15!white,
	coltitle=black,
	fonttitle=\bfseries,
	arc=3mm,
	boxrule=0.8pt,
	breakable
]
\small
\textit{The Output Format Requirements, Think-Act-Observe paradigm definition, and Phase Transition rules are identical to those in the Configuration Phase (Box~8). The following content replaces only the behavioral norms, selection principles, and checklist sections.}

\medskip\hrule\medskip

\textbf{Protocol Phase System Prompt}

You are an expert AI/CS Researcher. Your goal is to design the experimental protocol by populating the protocol queue with main experiments, ablation studies, and analysis experiments based on the provided research task and the datasets and baselines already selected during the configuration phase.

\medskip
\textbf{BEHAVIORAL NORMS IN PROTOCOL PHASE}

Follow the norms below strictly:
\begin{itemize}[nosep,leftmargin=12pt]
	\item \textbf{No Resource Filtering:} The datasets and baselines have already been selected during the configuration phase and will be provided to you. Your job is NOT to filter or screen them, but to assign appropriate datasets and baselines to each experiment.
	\item \textbf{Step-by-Step Design:} Design and populate the three parts (Main, Ablation, Analysis) in the protocol queue using the available tools (e.g., \texttt{manage\_main\_experiment\_step}, \texttt{manage\_sub\_experiment}, \texttt{write\_overall\_summary}).
	\item \textbf{Verification:} After each set of modifications, call \texttt{get\_protocal\_summary} or \texttt{read\_full\_protocal} to verify the current plan details. If you are not fully confident about the current state of the protocol after multiple changes, you MUST use these tools to confirm before proceeding.
	\item \textbf{Self-Review:} Periodically review the entire protocol design against the principles below. Self-evaluate and refine as needed.
	\item \textbf{Completion:} When you are CERTAIN the protocol is comprehensive, rigorous, and fully documented, output \texttt{<phase\_transition/>} to signal completion.
\end{itemize}

\medskip
\textbf{EXPERIMENT DESIGN REQUIREMENTS}

\textbf{1. Main Experiment (Core Claims Validation)}
\begin{itemize}[nosep,leftmargin=12pt]
	\item Build a complete, rigorous, and credible experimental pipeline.
	\item Must specify clear and detailed protocols for: data loading, data preprocessing, model inference, post-processing, and evaluation.
	\item Select appropriate datasets and baselines from the configuration queue.
	\item Choose evaluation metrics that align with the core research claims.
\end{itemize}

\textbf{2. Ablation Studies (Component Isolation)}
\begin{itemize}[nosep,leftmargin=12pt]
	\item Consists of multiple sub-experiments.
	\item Each sub-experiment must clearly identify the ablated component and ensure a fair, single-variable strategy.
	\item For each sub-experiment, specify the datasets, baselines, and metrics to be used.
	\item Ensure the ablation design is rigorous and convincingly validates the contribution of each key component.
\end{itemize}

\textbf{3. Analysis Experiments (Additional Insights)}
\begin{itemize}[nosep,leftmargin=12pt]
	\item Provide extra insights beyond main and ablation results (e.g., sensitivity analysis, hyperparameter analysis, efficiency analysis, data characteristic analysis).
	\item Identify analysis points that are genuinely valuable for understanding the method.
	\item Each sub-experiment should state the analysis objective, strategy, and the datasets/baselines/metrics used.
\end{itemize}

\medskip
\textbf{AVAILABLE TOOLS}

\{tool\_schemas\}
\end{tcolorbox}

\section{Case Study}

This section presents three representative cases from \textsc{SCOPE}: a success case where a baseline model produces a high-quality experimental plan aligned with ground truth; a failure case exposing systematic shortcomings such as hallucinated resources or missing ablation components; and a comparative case where OptED substantially improves upon a baseline, demonstrating both quantitative gains and transparency benefits of the stage-isolated, auditable workflow.

\subsection{Case 1: SE-Agent --- Successful Alignment with Ground Truth (Positive Example)}

This case originates from paper \textit{SE-Agent: Self-Evolution Trajectory Optimization in Multi-Step Reasoning with LLM-Based Agents}\cite{guo2026se}, which proposes a trajectory-level self-evolution framework for LLM-based code agents. SE-Agent iteratively refines multi-step agent interaction trajectories through three operators---Revision, Recombination, and Refinement---to increase search diversity, exploit cross-trajectory synergies, and avoid redundant reasoning. The ground-truth experimental design is tightly focused: it evaluates SE-Agent on SWE-bench Verified (a curated 500-instance benchmark of real GitHub issues), measures Pass@1 and Pass@5 across five LLM backbones, and compares against two baselines (SWE-Agent and the MCTS-based SWE-Search). Ablation studies isolate three components (w/o Revision, w/o Recombination, w/o All), and analysis experiments sweep cost-performance tradeoffs and examine overlap in resolved instances.

 GPT-5.2 under CoT+Search achieves a total score of 24/30 with no redline violation: Main Experiment (4/5), Ablation Studies (4/5), Analysis Experiments (4/5), Datasets (3/5), Baselines (4/5), and Metrics (5/5). The plan correctly reconstructs the end-to-end experimental paradigm---harness-based evaluation with a full SE-Agent evolution loop (initial trajectory pool $\rightarrow$ Revision $\rightarrow$ Recombination $\rightarrow$ Refinement with convergence)---and enhances rigor beyond the GT through explicit budget-matching, artifact logging, and lineage tracking. The Metrics dimension earns a perfect score by centering on the correct Pass@1/Pass@5 via the official harness. The ablation design isolates the two core components and contributes additional informative variants. On the Low-Level, the plan independently incorporates SWE-bench Full and Lite alongside Verified, and Agentless and AutoCodeRover alongside the required baselines. None of the GT's key resources are omitted; rather, the plan proposes a broader experimental scope that remains scientifically aligned with the research task and contributes to more comprehensive validation. These additional selections are well-motivated for the code-agent domain, and the resulting scores---3/5 and 4/5, both within the rubric's ``High'' and ``Excellent'' tiers---reflect that the core requirements are fully satisfied while the extensions differ from the specific reporting conventions of the original paper.

 This case is representative of a class of tasks where LLMs perform strongly: the research scope centers on a well-established benchmark with standardized evaluation protocols, the Low-Level configuration is straightforward, and the experimental logic is closely coupled to the described methodology. Under these conditions, the model's strength in high-level reasoning and protocol planning translates directly to quality output, while independently proposed resources that align with the task domain can further strengthen the experimental design rather than detract from it.

\begin{tcolorbox}[
	title={Box 10: Case 1 --- Research Task Overview(Input)},
	colback=blue!3!white,
	colframe=blue!30!black,
	colbacktitle=blue!15!white,
	coltitle=black,
	fonttitle=\bfseries,
	arc=3mm,
	boxrule=0.8pt,
	breakable
]
\small
\textbf{Research Summary.} SE-Agent is a self-evolution framework for LLM-based agents that optimizes multi-step interaction trajectories rather than treating them as independent rollouts. It iteratively improves a population of agent trajectories using three trajectory-level operators---Revision, Recombination, and Refinement---to increase search diversity, exploit cross-trajectory synergies, and avoid redundant reasoning. The framework is evaluated on SWE-bench Verified (500 real GitHub issues) by integrating as a plug-and-play module into an existing code-agent framework, improving Pass@1/Pass@5 across multiple LLMs.

\smallskip
\textbf{Research Questions.} (1) How can an LLM-based agent leverage past interaction trajectories to improve future multi-step reasoning and tool use? (2) How can genuinely diverse solution trajectories be generated beyond surface-level diversity from sampling? (3) How can cross-trajectory learning escape local optima and reduce redundant reasoning? (4) How can solution quality and diversity be balanced when selecting and evolving candidate trajectories?

\smallskip
\textbf{Contributions.} (i) Proposes SE-Agent, a trajectory-level self-evolution framework that iteratively revisits and improves pilot trajectories. (ii) Introduces three operators: Revision (reflection-driven single-trajectory improvement), Recombination (cross-trajectory synthesis via crossover/transfer/restructuring), and Refinement (multi-dimensional evaluation with quality--diversity-balanced selection). (iii) Demonstrates consistent gains on SWE-bench Verified across five LLM backbones compared with SWE-Agent and an MCTS-based SWE-Search baseline.

\smallskip
\textbf{Method Overview.} SE-Agent maintains a pool of candidate trajectories and iteratively evolves them. First, a diverse initial pool ($N=10$) is created via multi-planning exploration and mutation-based diversification. Then, in each evolution cycle: Revision reflects on each trajectory and produces targeted revisions; Recombination synthesizes new trajectories via crossover, transfer, and restructuring across trajectories; Refinement evaluates all candidates with a multi-dimensional reward (TaskCompletion $+$ ReasoningQuality $+$ Efficiency) and selects elite trajectories balancing reward and diversity. The loop terminates when improvement saturates or a preset iteration budget is reached.

\smallskip
\textbf{Constraints.} Open-source models run locally on NVIDIA A100 GPUs (80\,GB); closed-source models accessed via OpenAI and Anthropic APIs.
\end{tcolorbox}

\begin{tcolorbox}[
	title={Box 11: Case 1 --- Ground Truth Experimental Design Overview},
	colback=blue!3!white,
	colframe=blue!30!black,
	colbacktitle=blue!15!white,
	coltitle=black,
	fonttitle=\bfseries,
	arc=3mm,
	boxrule=0.8pt,
	breakable
]
\small
\textbf{Datasets and Metrics.} The evaluation uses SWE-bench Verified, a curated 500-instance benchmark of real-world GitHub issues where correctness is validated by developer-written unit tests via the official SWE-bench Docker harness. The primary metric is Pass@1 (percentage of issues resolved on the first attempt); the secondary metric is Pass@5 (resolved within five attempts).

\smallskip
\textbf{Baselines.} Two baselines are used: (1) SWE-Agent, a foundational agent-framework baseline for repository-level issue resolution; (2) SWE-Search, an MCTS-based baseline that combines search with iterative refinement. Both are evaluated across five LLM backbones: DeepSeek-V3-0324, Qwen-2.5-72B-Instruct, Llama-3.1-70B-Instruct, GPT-4o, and Claude-3.7-Sonnet.

\smallskip
\textbf{Main Experiment.} SE-Agent is configured with a default candidate trajectory count of 10. Initial trajectories are generated using five distinct planning strategies combined with mutation-based diversification. The evolution loop then applies Revision (reflection-driven single-trajectory improvement), Recombination (cross-trajectory synthesis via crossover, transfer, and restructuring), and Refinement (multi-dimensional reward-based evaluation with quality--diversity-balanced selection). The same prompt format is used across all evaluated models. Open-source models run on NVIDIA A100 GPUs (80\,GB); closed-source models are accessed via official APIs. Pass@1 and Pass@5 are reported on SWE-bench Verified and compared against the two baselines.

\smallskip
\textbf{Ablation Studies.} Three components are isolated: (1) w/o Revision---removing the Revision operator while retaining Recombination and Refinement; (2) w/o Recombination---removing the Recombination operator while retaining Revision and Refinement; (3) w/o All---removing all three trajectory optimization operators, corresponding to generating multiple trajectories without the SE-Agent evolution mechanism.

\smallskip
\textbf{Analysis Experiments.} Four analyses are conducted: (i) varying the number of candidate trajectories (5, 10, 15, 20) and measuring Pass@1; (ii) sweeping maximum API cost budget (0, 2, 4, 6, 8) and comparing Pass@1 against baselines; (iii) resolved-issue overlap analysis via Venn diagrams to identify unique vs.\ shared solves across frameworks; (iv) qualitative case study of trajectory diversity on a representative issue, comparing fix locations and demonstrating that SE-Agent produces root-cause patches rather than local quick-fixes.
\end{tcolorbox}

\begin{tcolorbox}[
	title={Box 12: Case 1 --- GPT-5.2 CoT+Search Output},
	colback=blue!3!white,
	colframe=blue!30!black,
	colbacktitle=blue!15!white,
	coltitle=black,
	fonttitle=\bfseries,
	arc=3mm,
	boxrule=0.8pt,
	breakable
]
\small
\textbf{Datasets.} (1) SWE-bench Verified (500 instances) as the primary evaluation benchmark; (2) SWE-bench Lite for development and ablation studies, with a documented 80/20 split for reproducible tuning; (3) SWE-bench Full (v1.x, 2,294 instances) for extended analyses and leaderboard comparison.

\smallskip
\textbf{Metrics.} Primary: Pass@1 (fraction of instances resolved on first attempt via the official harness). Secondary: Pass@5 (resolved within five candidates under fixed compute budget); Cost (total tokens, LLM calls, wall-clock time per instance); Tool-use efficiency (number of tool invocations and trajectory steps); Trajectory-level diversity (pairwise edit distance over action sequences, edited-file-set Jaccard similarity, diff similarity among patches); Cross-trajectory synergy (percentage of wins attributable to recombination vs.\ revision vs.\ initial pool).

\smallskip
\textbf{Baselines.} (1) SWE-agent, the foundational agent-framework baseline; (2) SWE-Search, an MCTS-based search + iterative refinement baseline; (3) Agentless, a localize$\rightarrow$repair$\rightarrow$validate workflow; (4) AutoCodeRover, a structure-aware autonomous issue-resolution system; (5) LingmaAgent / RepoUnderstander, a repository understanding and exploration agent. All are evaluated under matched compute budgets and identical sandbox constraints.

\smallskip
\textbf{Main Experiment (5 steps).} (1) Environment setup: download all SWE-bench datasets, pin harness version and Docker images, record prompt templates as versioned artifacts. (2) Baseline execution under matched budgets: two fairness regimes (fixed wall-clock budget per instance; fixed token/LLM-call budget), with full logging of tool calls, edits, and final patches. (3) SE-Agent integration: produce initial trajectory pool $T_0$ ($N=10$) via multi-planning prompts and mutation diversification; iterate Revision (reflection + targeted revision), Recombination (crossover/transfer/restructure), and Refinement (multi-dimensional reward with quality--diversity selection), logging per-trajectory rewards, diversity scores, and parent links. (4) Submission formatting: convert final patches (top-5 for Pass@5) into SWE-bench predictions format and run official harness evaluation with fixed settings. (5) Statistical reporting: paired bootstrap resampling over instance IDs for 95\% confidence intervals on Pass@1/Pass@5 differences; performance-vs-cost curves for each backbone and fairness regime; overlap analysis with operator attribution (initial vs.\ revision vs.\ recombination).

\smallskip
\textbf{Ablation Studies (6 variants).} \textit{Ablation-1 (w/o multi-planning):} fixed single prompt with no strategy variation for initial pool generation. \textit{Ablation-2 (w/o mutation diversification):} independently sampled seeds only, no controlled mutations. \textit{Ablation-3 (w/o Revision):} skip reflect-and-revise, reallocating revision budget to recombination for fairness. \textit{Ablation-4 (w/o Recombination):} skip crossover/transfer/restructure, matched via additional revised variants. \textit{Ablation-5 (w/o diversity selection):} rank and select purely by reward without dissimilarity constraint. \textit{Ablation-6 (recombination sub-operator isolation):} three separate runs with crossover-only, transfer-only, and restructure-only, tracking parent attribution per sub-operator.

\smallskip
\textbf{Analysis Experiments (5 studies).} \textit{Quality--diversity tradeoff:} grid-search diversity coefficient $w_{div} \in \{0.0,0.1,0.2,0.4,0.8\}$, plot Pass@1/Pass@5 vs.\ diversity and identify Pareto frontier. \textit{Pool-size sensitivity:} grid $N \in \{2,5,10,20\}$ and $k \in \{2,5,10\}$ with fixed total LLM calls, report marginal gain per additional candidate. \textit{Convergence dynamics:} per-iteration curves of best/mean Reward, diversity, and solved-status transitions; vary $\epsilon \in \{0,0.005,0.01,0.02\}$ and patience $\in \{1,2,3\}$, analyze recombination-driven escapes from revision-only stagnation. \textit{Efficiency and anytime behavior:} performance-vs-cost curves under increasing time budgets (5/10/20/30 min) and fixed token budgets; anytime curves via best-patch checkpoints. \textit{Unique-solve overlap and operator attribution:} solved-instance overlap sets across methods, with winning patches attributed to initial pool vs.\ revision vs.\ recombination via lineage tracking.
\end{tcolorbox}

\subsection{Case 2: SongBloom --- Missed Paradigm and Configuration Divergence (Negative Example)}

This case originates from paper \textit{SongBloom: Coherent Song Generation via Interleaved Autoregressive Sketching and Diffusion Refinement}\cite{yang2026songbloom}, which proposes an autoregressive diffusion framework for full-length lyric-to-song generation. SongBloom unifies semantic sketch generation and acoustic synthesis within an interleaved patch-wise architecture: a discrete sketch stream (quantized MuQ embeddings) is generated autoregressively in patches, and for each patch a hidden vector conditions a non-autoregressive diffusion transformer that synthesizes continuous acoustic latents via rectified flow matching. The interleaving allows bidirectional information flow---prior acoustic context informs subsequent sketch planning, while the sketch guides acoustic refinement within each patch. Training is end-to-end with a joint objective (cross-entropy + flow matching, $\lambda=0.1$) on a large-scale proprietary Chinese+English song corpus ($\sim$100K hours). Evaluation spans lyric alignment (PER), style consistency (MCC), audio fidelity (FAD), structural correctness (SER), Audiobox-Aesthetic scores, detailed MOS, and RTF against 8 commercial and open-source baselines.

Qwen3-Max under CoT+Search achieves a total score of 9/30 with no redline violation: Main Experiment (1/5), Ablation Studies (2/5), Analysis Experiments (3/5), Datasets (1/5), Baselines (1/5), and Metrics (1/5). The central failure is a fundamental misalignment with the paper's experimental paradigm. The plan omits the core training-and-inference pipeline that substantiates SongBloom's claims---the dataset cleaning/annotation pipeline, sketch-token and continuous-latent representation extraction, explicit model configuration (two sizes), and, most critically, the joint training setup (LM+flow objective, diffusion-time sampling, CFG masking during training, optimizer/schedule/steps). Instead of reproducing the GT's end-to-end training-based validation, the plan proposes an evaluation-centric workflow that treats models as pre-trained black boxes, measuring objective metrics and MOS on a self-constructed benchmark. This constitutes a category-level departure from the paper's experimental logic: designing an experiment to evaluate a lyric-to-song system, rather than designing an experiment to validate SongBloom's architectural claims.

On the Low-Level, the configuration divergence is systematic and mutually reinforcing. The GT's key dataset---a $\sim$100K-hour proprietary bilingual song corpus with a multi-stage cleaning/alignment/structure pipeline---is replaced by a 200-song publicly-sourced evaluation benchmark with no training-scale counterpart, making it impossible to reproduce training-dependent claims. The GT's 8 baselines, spanning commercial systems (Suno, Udio, Haimian, Mureka-O1) and open-source methods (ACE-step, YuE, DiffRhythm, SongEditor), are replaced by three different baselines (MusicGen-Melody, a DiffSinger+MusicGen pipeline, and a non-interleaved internal variant) that are directionally related but do not align with the GT comparison set. The primary metric shifts from PER (phoneme error rate on separated vocals, measuring recognition/alignment correctness) to PAE (phoneme alignment error via forced alignment, measuring timing deviation), and several GT-specific metrics (MCC, SER, Audiobox-Aesthetic sub-scores, detailed MOS categories) are omitted. Cumulatively, these configuration choices mean that even if the proposed experiment were executed successfully, its results would not be directly comparable to the original paper's findings.

Despite the low overall score, the plan is not without merit. The Analysis dimension earns 3/5 through well-structured sensitivity studies on patch size (sweeping $\{8,16,32\}$, plotting RTF-vs-FAD and PAE-vs-Structure F1 Pareto frontiers), diffusion step efficiency ($K\in\{18,36,72\}$ with >95\% quality threshold), and style prompt length robustness---directly probing the paper's core quality--runtime tradeoff. The ablation design (2/5) targets meaningful mechanisms (removing acoustic context tokens, altering the per-patch hidden vector pathway, varying CFG) that are conceptually aligned with the paper's architecture. These strengths, however, cannot compensate for the absence of the training paradigm that defines the paper's contribution.

This case illustrates a pattern common among lower-performing models on \textsc{SCOPE}: when a paper's experimental paradigm involves large-scale training with proprietary data and a tightly specified pipeline (representation extraction $\rightarrow$ joint optimization $\rightarrow$ inference), models tend to default to an evaluation-only plan that treats the proposed method as an off-the-shelf system rather than a contribution to be validated through reproduction. The High-Level design collapses to ``benchmark the model'' because the training-dependent validation logic is not reconstructed; the Low-Level configuration then drifts because the substitute evaluation setting has different requirements. In contrast to Case~1 where the task centers on a standardized benchmark with well-defined evaluation protocols, SongBloom's experimental claims are inseparable from its training pipeline, creating a structural challenge that Qwen3-Max fails to overcome. This failure mode suggests that benchmark designs relying on training-based validation may require explicit scaffolding---e.g., prompts that surface the necessity of data preparation, representation extraction, and training configuration as integral experimental operations---to prevent models from defaulting to evaluation-only reasoning.

\begin{tcolorbox}[
	title={Box 13: Case 2 --- Research Task Overview (Input)},
	colback=red!3!white,
	colframe=red!30!black,
	colbacktitle=red!15!white,
	coltitle=black,
	fonttitle=\bfseries,
	arc=3mm,
	boxrule=0.8pt,
	breakable
]
\small
\textbf{Research Summary.} SongBloom is an autoregressive diffusion framework for full-length lyric-to-song generation (up to 150 seconds, 48\,kHz stereo). It interleaves autoregressive generation of discrete semantic sketch tokens (from quantized MuQ SSL embeddings) with non-autoregressive diffusion refinement of continuous acoustic latents (from a waveform VAE), both segmented into fixed-size patches. This interleaving enables bidirectional information flow: previous acoustic context helps sketch planning, while the sketch guides acoustic synthesis within each patch. Training uses a joint objective (sketch-token cross-entropy + rectified flow matching, $\lambda=0.1$) with classifier-free guidance masking, on a large-scale proprietary Chinese+English song corpus.

\smallskip
\textbf{Research Questions.} (1) How can a model generate full-length songs that remain structurally coherent while maintaining high-fidelity audio and correct lyric alignment? (2) How can semantic planning and acoustic refinement be unified in a single framework without requiring the full semantic sequence before acoustic generation? (3) Can interleaving semantic and acoustic patches improve long-form consistency by allowing acoustic context to inform subsequent semantic planning? (4) How do patch size and diffusion inference steps affect the trade-off among alignment, musicality, fidelity, and runtime?

\smallskip
\textbf{Contributions.} (i) An autoregressive diffusion-based lyric-to-song generation framework producing full songs from structured lyrics and a 10-second reference audio prompt. (ii) An interleaved patch-wise generation paradigm alternating sketch-token prediction and acoustic-latent diffusion refinement, enabling bidirectional context exchange. (iii) A unified training setup combining sketch-token cross-entropy loss with rectified flow-matching diffusion loss, with gradients flowing from diffusion to the sketch stage via a per-patch hidden vector. (iv) Empirical results showing improved performance vs.\ open-source baselines and competitive comparisons with commercial song generation platforms, alongside favorable real-time factor among autoregressive baselines.

\smallskip
\textbf{Method Overview.} Lyrics are augmented with structure flags, normalized, and converted to phonemes. A discrete sketch stream is generated autoregressively in patches (patch size $P=16$, 0.64\,s); for each patch the model outputs a hidden vector $h_i$ that conditions a non-autoregressive diffusion transformer synthesizing continuous acoustic latents via rectified flow matching with Euler ODE solver (36 steps). The sketch generator conditions on prior sketch tokens and compressed acoustic context from previous patches; diffusion conditions on current sketch tokens, $h_i$, and previous-patch latents. CFG coefficient 1.5 is applied at inference. Two model sizes: SongBloom-tiny (AR 16L, diffusion 8L, max 60\,s) and SongBloom-full (AR 24L, diffusion 12L, max 150\,s). Training: AdamW ($lr=1e^{-4}$), cosine schedule with 2000 warm-up steps, batch size 128, 150K steps, 16 A100 GPUs, $\sim$1 week.

\smallskip
\textbf{Constraints.} Training requires 16$\times$A100 GPUs for $\sim$1 week per model run. Training data is proprietary (not publicly available). Models are for academic research use only; weights are not released.
\end{tcolorbox}

\begin{tcolorbox}[
	title={Box 14: Case 2 --- Ground Truth Experimental Design Overview},
	colback=red!3!white,
	colframe=red!30!black,
	colbacktitle=red!15!white,
	coltitle=black,
	fonttitle=\bfseries,
	arc=3mm,
	boxrule=0.8pt,
	breakable
]
\small
\textbf{Datasets and Metrics.} The key dataset is a large-scale proprietary Chinese+English song corpus ($\sim$100K hours) with a multi-stage cleaning/annotation pipeline: Demucs vocal/instrumental separation, WhisperX lyric-audio alignment filtering, lyric time boundary refinement, and structure extraction via audio structure analyzer. Lyrics are preprocessed with vocal-based structure flags (verse/chorus) and accompaniment-based flags (intro/outro), normalized, and converted to phonemes. Sketch tokens are extracted via MuQ+VQ (codebook size 16384, 25\,Hz); acoustic latents via a stable-audio-vae-based autoencoder at 25\,Hz. The primary metric is Phoneme Error Rate (PER) on separated vocals. Secondary metrics: MuLan Cycle Consistency (MCC), Fr\'{e}chet Audio Distance (FAD), Structural Error Rate (SER) via DTW-based structure error-duration proportion, Audiobox-Aesthetic sub-scores (CE, CU, PC, PQ), detailed MOS (MUSV, MUSA, QLTV, QLTA, CRR, CST when applicable), and Real-Time Factor (RTF).

\smallskip
\textbf{Baselines.} Eight systems spanning commercial and open-source categories: Suno-v4.5, Udio-v1.5, Haimian, Mureka-O1, ACE-step, YuE-7B, DiffRhythm-full, and SongEditor. Prompt formats vary by system (text-only, 10-second audio, or 30-second audio as supported).

\smallskip
\textbf{Main Experiment (7 steps).} (1) Dataset preparation: apply Demucs, WhisperX, structure analyzer; preprocess lyrics with structure flags and phoneme conversion. (2) Representation extraction: compute MuQ+VQ sketch tokens and VAE acoustic latents synchronized at 25\,Hz. (3) Model configuration: SongBloom-tiny (AR 16L, diffusion 8L) and SongBloom-full (AR 24L, diffusion 12L), patch size $P=16$, 24 attention heads, hidden dim 1536, two-layer convolutional acoustic encoder. (4) Joint training: AdamW ($lr=1e^{-4}$), cosine schedule (2000 warm-up), batch size 128, $\sim$150K steps, DeepSpeed on 16 A100 GPUs ($\sim$1 week). Loss: $\mathcal{L}=\mathcal{L}_{LM}+0.1\mathcal{L}_{flow}$ with logit-normal diffusion time sampling. CFG training via conditioning signal masking (hidden vector + sketch tokens as unified condition, plus sketch-token-only masks). (5) Inference: top-$k$=200 sampling, temperature 0.9, Euler ODE solver (36 steps), CFG coefficient 1.5. (6) Fine-tuning variant: SongBloom-full-ft, 1000 additional steps on synthesized alternating verse/chorus structured data. (7) Evaluation: PER (via separated vocals), MCC, FAD, SER (DTW-based), Audiobox-Aesthetic, MOS ($\geq$10 musically experienced participants), RTF.

\smallskip
\textbf{Ablation Studies (4 studies, SongBloom-tiny).} (1) Diffusion condition inputs: factorize conditioning set as $\{H$: hidden vector, $C$: acoustic context, $S$: sketch tokens$\}$ and compare $H+C+S$ vs.\ $H+C$ vs.\ $H$ (all with sketch enabled) plus a no-sketch setting (LM produces only hidden vectors, diffusion uses $H+C$). (2) Patch size study: vary $P$ and compare against the $+\infty$ extreme (decoupled two-stage generation), measuring aesthetic scores and sketch accuracy. (3) Diffusion inference steps: vary Euler solver step count, report RTF-vs-quality trade-off. (4) Sketch representation choice: compare no sketch vs.\ pitch vs.\ pitch+chroma vs.\ SSL (MuQ), measuring PER, FAD, aesthetic scores.

\smallskip
\textbf{Analysis Experiments (2 studies).} (i) Inference time-complexity comparison vs.\ decoupled two-stage model, deriving asymptotic terms under KV-caching for variables $L_1$ (LM layers), $L_2$ (diffusion layers), $T$ (total frames), $P$ (patch size), $S$ (diffusion steps). (ii) Model scaling and fine-tuning effect: compare SongBloom-tiny vs.\ SongBloom-full vs.\ SongBloom-full-ft on PER, MCC, FAD, SER, Audiobox-Aesthetic, MOS, RTF.
\end{tcolorbox}

\begin{tcolorbox}[
	title={Box 15: Case 2 --- Qwen3-Max CoT+Search Output},
	colback=red!3!white,
	colframe=red!30!black,
	colbacktitle=red!15!white,
	coltitle=black,
	fonttitle=\bfseries,
	arc=3mm,
	boxrule=0.8pt,
	breakable
]
\small
\textbf{Datasets.} A single self-constructed evaluation benchmark (200 songs) from royalty-free sources (ccMixter, FreePD, Open Music Archive) with human-annotated structure flags (verse, chorus, bridge, intro, outro; Fleiss' $\kappa>0.85$), 10-second style prompts, and ground-truth audio ($\leq$150\,s, 48\,kHz stereo). Split: 100 validation / 100 test. No training-scale dataset is specified.

\smallskip
\textbf{Metrics.} Primary: Phoneme Alignment Error (PAE) via Montreal Forced Aligner v2.0. Secondary: Fr\'{e}chet Audio Distance (FAD) via VGGish embeddings; Structure Label F1-score via madmom pattern tracking; Real-Time Factor (RTF) on A100 80\,GB; Mean Opinion Score (MOS) for Overall Quality, Lyric Intelligibility, Style Consistency, Structural Coherence, and Naturalness (20 raters, randomized blinded protocol).

\smallskip
\textbf{Baselines.} (1) MusicGen-Melody (3.3B), an open-source text-and-melody-conditioned music generation model; (2) Pipeline Baseline (DiffSinger + MusicGen), a two-stage pipeline with DiffSinger vocal generation and MusicGen accompaniment mixed at fixed ratio; (3) Non-Interleaved SongBloom, an internal variant generating the full sketch sequence before acoustic diffusion. No commercial baselines are included.

\smallskip
\textbf{Main Experiment (4 steps).} (1) Preprocessing: load SongBloom-Eval test set, preprocess lyrics (structure flags, normalize, phonemize with espeak-ng), extract 10-second style prompts at 48\,kHz. (2) Generate songs from all baselines under fixed configurations: SongBloom (patch size 16, 36 diffusion steps, CFG 1.5), MusicGen-Melody (chunked 30-second generation with 5-second overlap blending), Pipeline Baseline, Non-Interleaved SongBloom; normalize all outputs to $-3$\,dB peak. (3) Compute objective metrics: PAE, FAD, Structure F1, RTF. (4) Human MOS study: 20 raters, 10 randomly selected songs per model, 1--5 scale across 5 dimensions.

\smallskip
\textbf{Ablation Studies (3 studies, SongBloom-tiny).} (1) Interleaving mechanism: remove acoustic context tokens from the LM stream, retrain with acoustic encoder disabled; evaluate Structure F1 and PAE at patch boundaries. (2) Per-patch hidden vector: replace learned $h_i$ with mean-pooled sketch token embeddings; evaluate FAD and style consistency FAD. (3) CFG ablation: run SongBloom-full inference with CFG scale 1.0 vs.\ 1.5; compare MOS scores via paired $t$-tests.

\smallskip
\textbf{Analysis Experiments (3 studies).} (i) Patch size sensitivity: grid search $P\in\{8,16,32\}$ (0.32\,s, 0.64\,s, 1.28\,s), plot Pareto frontiers of RTF vs.\ FAD and PAE vs.\ Structure F1. (ii) Diffusion step efficiency: vary $K\in\{18,36,72\}$ steps, determine minimal steps achieving >95\% of max FAD, measure RTF impact. (iii) Style prompt length robustness: truncate prompts to 2\,s/5\,s/10\,s/20\,s, compute Style Consistency FAD between truncated prompt and first $N$ seconds of generation, analyze degradation curve.
\end{tcolorbox}

\subsection{Case 3: Animate-X --- OptED-Driven Improvement over Baseline (Framework Advantage)}

This case originates from paper \textit{Animate-X: Universal Character Image Animation with Enhanced Motion Representation}\cite{tan2024animate}, which proposes a universal character image animation framework that generalizes pose-guided human image animation to anthropomorphic characters through a Pose Indicator comprising an Implicit Pose Indicator (IPI, extracting motion gist from driving-video CLIP features via pose-guided cross-attention) and an Explicit Pose Indicator (EPI, simulating reference/pose misalignment via anchor realignment and rescaling/part-edit operations during training). The GT experimental design is methodologically rich: 8 fine-grained ablation variants within the Pose Indicator and 4 targeted analysis experiments---making it a strong test of whether an agentic workflow can recover a complex, component-level validation strategy.

The DeepSeek V3.2 baseline (CoT+Search) achieves 11/30 under GPT evaluation (Main 2, Ablation 2, Analysis 3, Datasets 1, Baselines 1, Metrics 2). On the High-Level, the baseline proposes only 3 coarse ablations (IPI removal, EPI removal, temporal module swap), omitting the GT's dissection of IPI query design and EPI transformation sub-components, and its analyses miss the A2Bench difficulty-level breakdown. On the Low-Level, the baseline selects datasets (TaiChi, UBCFashion, Human3.6M, MGIF, A2Bench) and baselines (MagicAnimate, AnimateAnyone, FOMM, plus generic ControlNet-for-video) that deviate from the GT, omitting Moore-AnimateAnyone, ControlNeXt, MusePose, and UniAnimate.

When processed through OptED, the framework produces a substantially strengthened design. On the High-Level, the Protocol Queue records 5 ablation studies---including a systematic EPI component-wise breakdown isolating anchor realignment, pose rescaling, and part editing that directly mirrors the GT's strategy---and 5 analysis experiments covering controlled pose perturbation robustness, anthropomorphic character type breakdown (validating the ``universal'' claim), inference efficiency profiling, hyperparameter sensitivity, and cross-domain generalization. The character type breakdown provides structured evidence for the generalization hypothesis that the baseline's plan does not address. On the Low-Level, the Configuration Queue (CQ) records a more deliberate selection process. For datasets, the DeepSeek baseline includes TaiChi based on general relevance to human motion transfer; OptED explicitly evaluates TaiChi and excludes it with the rationale: ``Less commonly used in recent diffusion-based animation works. Standard benchmarks (TikTok, UBCFashion) provide sufficient coverage for fair comparison. TaiChi does not directly test anthropomorphic generalization which is the key novelty.'' For baselines, OptED surveys 9 candidates through a structured search phase and selects 7 (AnimateAnyone, MagicAnimate, DisCo, Champ, MimicMotion, VividPose, TCAN) spanning distinct methodological families, while excluding PoseAnimate as a training-free paradigm mismatch. Each CQ entry carries source attribution, reported performance, and task-grounded justification.

Beyond output quality, OptED's workflow architecture provides a layer of transparency absent from the baseline. The CQ (24 timestamped versions), Protocol Queue, and reading list (14 candidate papers with target roles and resolution status) serve as explicit, auditable records of every decision. The checkpoint system preserves the full action-reasoning trajectory, enabling post-hoc inspection of why datasets were excluded, which performance data informed baseline selection, and how the ablation design evolved. While a single JSON output from a baseline model presents only the final answer, OptED's CQ--PQ--checkpoint trail allows a human researcher to verify that each decision is evidence-based and scientifically defensible---a prerequisite for responsible deployment when ground truth is unavailable.

This case illustrates three interacting contributions of OptED. First, stage isolation (Configuration $\rightarrow$ Protocol $\rightarrow$ Reporting) enables deeper exploration: the search phase systematically surveys candidates rather than accepting the first plausible match, and the protocol phase constructs a more complete and granular experiment matrix. Second, the structured CQ enforces explicit reasoning about resource inclusion and exclusion, transforming implicit selections into documented, contestable decisions. Third, the full workflow trail renders the model an auditable reasoning partner rather than an opaque recommendation engine. The cumulative effect spans both High-Level design (5 vs.\ 3 ablations, 5 vs.\ 3 analyses, with finer granularity and closer GT alignment) and Low-Level configuration (TaiChi excluded with reasoning, PoseAnimate excluded with paradigm justification, 7 methodologically diverse baselines), even though configuration scores remain constrained by the inherent difficulty of matching proprietary training data under the benchmark's scoring rubric.

\begin{tcolorbox}[
	title={Box 16: Case 3 --- Research Task Overview (Input)},
	colback=green!3!white,
	colframe=green!30!black,
	colbacktitle=green!15!white,
	coltitle=black,
	fonttitle=\bfseries,
	arc=3mm,
	boxrule=0.8pt,
	breakable
]
\small
\textbf{Research Summary.} Animate-X is a universal character image animation framework based on latent diffusion models that generalizes from human-only training to diverse character types, especially anthropomorphic characters with non-human body structures. The key innovation is a Pose Indicator with two components: an Implicit Pose Indicator (IPI) that extracts motion gist and temporal relations from driving-video CLIP visual features using a query-based cross-attention extractor guided by pose keypoints plus a learnable query; and an Explicit Pose Indicator (EPI) that improves robustness to reference/pose misalignment by simulating body-shape mismatches during training via anchor pose realignment and rescaling/part-edit operations. The denoiser uses a 3D-UNet with Spatial Attention (identity-motion fusion), Motion Attention (IPI injection), and a Mamba-based temporal module. The paper also introduces A2Bench, a 500-pair animated anthropomorphic benchmark with a manually screened 100-video subset for quantitative evaluation.

\smallskip
\textbf{Research Questions.} (1) How can a pose-guided image animation system generalize beyond humans to anthropomorphic characters with different body structures and missing/extra parts? (2) How can motion representation be enhanced to include both explicit pose structure and implicit motion patterns from the driving video? (3) How can training be modified to handle reference/pose misalignment at inference without training on anthropomorphic paired data? (4) How can anthropomorphic character animation be evaluated quantitatively when pose extraction is unreliable on non-human characters?

\smallskip
\textbf{Contributions.} (i) A latent-diffusion-based animation framework (Animate-X) that generalizes from human training videos to diverse character types without strict pose alignment. (ii) Pose Indicator comprising IPI (CLIP-video-feature-driven motion gist extraction with pose-guided and learnable queries) and EPI (training-time pose transformations via anchor realignment and rescaling/part-edit operations). (iii) A2Bench: a 500-pair animated anthropomorphic benchmark with diverse styles and a manually screened 100-video quantitative subset. (iv) Extensive comparisons and ablations showing improved identity preservation, motion consistency, and robustness to misaligned pose inputs.

\smallskip
\textbf{Method Overview.} A reference image is encoded by a CLIP image encoder (appearance feature) and a VAE encoder (reference latent). The driving video provides CLIP visual features (used by IPI to extract motion gist) and DWPose keypoints (processed by EPI with training-time transformations). The 3D-UNet denoiser uses Spatial Attention to fuse identity/motion, Motion Attention to inject IPI features, and Mamba-based temporal modeling for consistency. Trained with DDPM (1000 steps), AdamW ($lr_{IPI}=5e^{-7}$, $lr_{other}=5e^{-5}$), EPI probability $\lambda>0.98$, 8 A100 GPUs. Inference uses DDIM (50 steps). Training data: $\sim$9,000 internet human dance videos. Evaluation: TikTok (10 videos), Fashion (100 videos), A2Bench (100-video screened subset, self-driven and rescaled-pose settings).

\smallskip
\textbf{Constraints.} Training requires 8$\times$A100 GPUs. Training data is internally collected (not publicly released). The method generates videos rather than real-time streams due to iterative diffusion sampling.
\end{tcolorbox}

\begin{tcolorbox}[
	title={Box 17: Case 3 --- DeepSeek V3.2 CoT+Search Baseline Output},
	colback=green!3!white,
	colframe=green!30!black,
	colbacktitle=green!15!white,
	coltitle=black,
	fonttitle=\bfseries,
	arc=3mm,
	boxrule=0.8pt,
	breakable
]
\small
\textbf{Datasets.} (1) TaiChi Dataset: large-scale TaiChi movement videos with pose annotations ($>$10,000 sequences, 70/15/15 split). (2) UBC Fashion Dataset: pose-guided person image generation in fashion (thousands of pairs, 80/10/10 split). (3) Human3.6M: 3D human motion capture with multi-view video and accurate 3D pose annotations. (4) MGIF Dataset: GIF-based motion transfer dataset with pose-guided animation. (5) A2Bench (proposed): 500 anthropomorphic image-video pairs, with an additional 200-pair test set and 300-pair training/validation split for development.

\smallskip
\textbf{Metrics.} Primary: FVD (I3D backbone). Secondary: SSIM, PSNR, LPIPS (frame-level); FID-VID (video-level); CLIP Similarity and CSIM (identity preservation); Pose Distance Error (keypoint-level accuracy); User Study (MOS 1--5 for identity, temporal, visual quality, 20 evaluators, Fleiss' $\kappa$); Efficiency metrics (inference time, GPU memory, parameters).

\smallskip
\textbf{Baselines.} (1) MagicAnimate: appearance encoder + DensePose conditioning; (2) AnimateAnyone: ReferenceNet + Pose Guider + temporal attention; (3) FOMM: first-order motion model for image animation; (4) SVD with pose-based ControlNet: Stable Video Diffusion adapted with pose ControlNet conditioning.

\smallskip
\textbf{Main Experiment (5 steps).} (1) Data preprocessing: frame extraction, pose estimation (DWPose), resizing and normalization at 256$\times$256 (lower resolution than GT's 768$\times$512). (2) Diffusion training: 16-frame clips, DDPM with 1000 steps, AdamW ($lr=1e^{-4}$), batch size 4 per GPU, $\sim$50K iterations on 4 A100 GPUs. (3) DDIM inference with 50 steps. (4) Quantitative evaluation: compute FVD, SSIM, LPIPS, CSIM, and Pose Distance Error on test splits. (5) User study: 20 evaluators, 1--5 MOS scale, side-by-side comparisons with randomized ordering, Fleiss' $\kappa$ for inter-rater agreement.

\smallskip
\textbf{Ablation Studies (3 studies).} (1) w/o IPI: remove IPI module, replace with average pooling of CLIP features. (2) w/o EPI: disable all EPI pose transformations during training. (3) Temporal module swap: replace Mamba-based temporal module with standard transformer temporal attention.

\smallskip
\textbf{Analysis Experiments (3 studies).} (i) Hyperparameter sensitivity: sweep $\lambda$ (EPI transformation probability) $\in\{0.5,0.7,0.9,0.98\}$, number of diffusion sampling steps $\in\{10,25,50,100\}$, number of input frames $\in\{8,16,32\}$. (ii) Efficiency and resource analysis: parameters, inference time per frame, peak GPU memory, FLOPs for all methods. (iii) Robustness to pose noise: add Gaussian noise to DWPose keypoints at $\sigma\in\{2,5,10\}$ pixels, measure FVD and CSIM at each noise level.
\end{tcolorbox}

\begin{tcolorbox}[
	title={Box 18: Case 3 --- OptED Framework Output (Configuration Queue + Protocol Queue Highlights)},
	colback=green!3!white,
	colframe=green!30!black,
	colbacktitle=green!15!white,
	coltitle=black,
	fonttitle=\bfseries,
	arc=3mm,
	boxrule=0.8pt,
	breakable
]
\small
\textbf{Configuration Queue (CQ, 24 versions).} The CQ records structured dataset and baseline decisions with status tracking (selected/candidate/excluded), source attribution, performance data, and rationale for each entry.

\smallskip
\textit{Datasets --- Selected (3):} (1) \textbf{TikTok Dataset}: $\sim$350 single-person dance videos, primary human animation benchmark used by all major baselines. Essential for fair comparison with prior work. (2) \textbf{UBCFashion Video Dataset}: 500 train / 100 test fashion videos with clean backgrounds. Provides domain diversity from TikTok; used by AnimateAnyone, VividPose, and Champ. (3) \textbf{A2Bench}: 500 anthropomorphic image-video pairs (100 manually screened for quantitative evaluation). Core benchmark for the paper's generalization claim. \textit{Datasets --- Excluded (1):} \textbf{TaiChi Dataset}: evaluated and excluded. Rationale: ``Less commonly used in recent diffusion-based animation works. Standard benchmarks provide sufficient coverage for fair comparison. TaiChi does not directly test anthropomorphic generalization which is the key novelty.''

\smallskip
\textit{Baselines --- Selected (7):} (1) \textbf{AnimateAnyone}: ReferenceNet family; TikTok SSIM 0.718, FVD 171.9. (2) \textbf{MagicAnimate}: appearance encoder + temporal attention family; TikTok FVD 179.07. (3) \textbf{DisCo}: disentangled control family; TikTok FVD 229.66. (4) \textbf{Champ}: SMPL-based 3D guidance family; TikTok FVD 160.82. (5) \textbf{MimicMotion}: SVD-based family; TikTok FVD 594. (6) \textbf{VividPose}: multi-condition fusion family; TikTok FVD 152.97. (7) \textbf{TCAN}: temporal consistency family; TikTok FVD 154.84. \textit{Baselines --- Excluded (1):} \textbf{PoseAnimate}: evaluated and excluded. Rationale: ``Training-free zero-shot approach represents a fundamentally different paradigm from Animate-X's training-based approach. Including it would dilute the comparison focus. The selected training-based baselines provide more directly comparable results.''

\smallskip
\textbf{Protocol Queue (PQ).} The PQ records the full experimental protocol with structured experiment entries.

\smallskip
\textit{Main Experiment (5 steps).} (1) Data preparation: load TikTok (768$\times$768, 8 FPS, DWPose), UBCFashion (512$\times$768), A2Bench (100-video screened subset); extract reference images and pose sequences. (2) Animate-X inference: SD v1.5 backbone, 3D-UNet with spatial/motion attention, Mamba temporal module, full IPI+EPI, DDIM 50 steps, 16-frame clips. (3) Baseline inference: all 7 baselines under matched settings. (4) Evaluation: SSIM, PSNR, LPIPS (frame-level); FVD, FID-VID (video-level, 16-frame clips); CSIM (identity preservation); A2Bench cross-identity uses FVD/FID-VID distribution metrics + user study (20 evaluators, 1--5 Likert, 3 criteria). (5) Reporting: per-dataset tables with best/second-best bolded, paired $t$-tests with $p<0.05$, inference time/GPU memory, qualitative frames at key timesteps, user study Fleiss' $\kappa$ and 95\% CIs.

\smallskip
\textit{Ablation Studies (5 studies).} (1) IPI motion gist: replace query-based cross-attention IPI with simple average pooling of CLIP features; evaluate on TikTok, UBCFashion, A2Bench. (2) EPI pose transformation: disable all EPI transformations (anchor realignment, rescaling, part editing); focus on A2Bench cross-identity where misalignment is most severe. (3) Pose-guided query: use only learnable query without pose keypoint guidance in IPI cross-attention. (4) Mamba vs.\ temporal attention: replace Mamba with standard transformer temporal attention; measure FVD and inference time. (5) EPI component-wise breakdown: separately ablate anchor realignment, pose rescaling, and part editing on A2Bench cross-identity to identify the dominant robustness mechanism.

\smallskip
\textit{Analysis Experiments (5 studies).} (i) Pose misalignment robustness: controlled perturbations (spatial translation $\pm$5--20\%, scaling 0.8--1.2$\times$, dropout 10--30\%) on TikTok and A2Bench; degradation curves for Animate-X vs.\ w/o EPI vs.\ AnimateAnyone/MagicAnimate/TCAN. (ii) Character type breakdown: classify A2Bench into humanoid/quadrupedal/winged/limb-missing/extra-limb categories; per-category FVD and CSIM for Animate-X vs.\ top-3 baselines. (iii) Efficiency comparison: parameters, inference time, GPU memory, FLOPs for all 8 methods; separate IPI/EPI overhead profiling. (iv) Hyperparameter sensitivity: DDIM steps $\{10,25,50,100\}$, IPI query dim $\{64,128,256,512\}$, Mamba expansion $\{1,2,4,8\}$, EPI rescaling $\{\pm10\%,\pm20\%,\pm30\%\}$. (v) Cross-domain generalization: qualitative evaluation on real-world non-human videos (animals, robots, mascots) as a complement to synthetic A2Bench.

\smallskip
\textbf{Workflow Transparency.} The reading list records 14 candidate papers with target roles (baseline/dataset) and resolution status. The CQ maintains version 24 with timestamps, enabling full rollback and audit. Each CQ entry carries a rationale field linking the decision to task-specific scientific considerations. The checkpoint system preserves the complete trajectory of search, configuration, and protocol construction actions, making every decision---from dataset exclusion to baseline family classification---retrospectively inspectable.
\end{tcolorbox}

\section{Limitation}

We discuss several dimensions of scope and boundary conditions that situate the contributions of \textsc{SCOPE} and outline directions for future work.

\textbf{Domain Scope.} \textsc{SCOPE} targets the experimental design workflow within computer science and AI research, drawing its task corpus from ICML, ICLR, and NeurIPS. The experimental paradigms captured in these venues---typically involving benchmark-driven evaluation of computational methods---differ substantially from those in the natural sciences. Research in biology, chemistry, and physics often involves physical experimentation, wet-lab protocols, equipment constraints, and safety considerations that lie outside the scope of the current benchmark. As AI systems increasingly support interdisciplinary scientific discovery, extending the benchmark to cover natural science experimental design workflows, with their distinct validation logic and resource constraints, represents an important direction for future investigation.

\textbf{Pipeline Stage.} \textsc{SCOPE} evaluates a specific, cognitively demanding stage of the research process: generating a complete experimental plan given only a research context and methodology description. It does not assess the subsequent stages of experimental execution, result analysis, or iterative refinement. In real-world research, experimental design is rarely a one-shot activity; it proceeds through feedback loops where preliminary results reveal unanticipated behaviors, prompting revisions to the experimental protocol, adjustments to hyperparameter ranges, or even reformulation of the research questions themselves. The static design-to-evaluation paradigm of \textsc{SCOPE}, while enabling rigorous and reproducible assessment, abstracts away this iterative dynamic. Future extensions could incorporate multi-turn experimental planning scenarios where models receive simulated experimental outcomes and must adapt their designs accordingly.

\textbf{Domain Distribution.} The domain coverage of \textsc{SCOPE} reflects the relative prevalence of research topics at the three source venues, with foundation models, generative models, and multimodal applications being the most represented areas. This distribution is a deliberate consequence of our curation strategy: we select papers based on community impact signals (GitHub stars and forks) to ensure that each task is grounded in a well-validated, technically substantive piece of work. High-impact papers naturally concentrate in active, fast-moving research areas, which leads to a domain distribution that mirrors the current landscape of the field rather than a uniform sampling across all areas. While this prioritization trades off representational balance for data quality, it means that performance estimates on \textsc{SCOPE} are most reliable for the mainstream research domains where the benchmark has the densest coverage. For domains with fewer tasks (e.g., learning theory, causal reasoning), per-domain results should be interpreted with appropriate caution.

\textbf{Extraction Fidelity.} The structured task representations are produced through an LLM-based extraction pipeline with multi-model verification and iterative quality refinement. Despite these safeguards, extraction is an inherently lossy process: nuances in the original paper's exposition, implicit assumptions, and domain-specific conventions may not survive translation into the standardized schema. The quality assessment threshold (35/50, 70\%) ensures a baseline level of fidelity, but does not guarantee completeness. We mitigate this concern through the multi-model extraction strategy, which provides cross-model agreement signals as auxiliary indicators of extraction reliability, and through the full provenance trail that enables downstream verification against the source paper.

\section{Impact Statement}

\textsc{SCOPE} aims to advance the systematic study of autonomous experimental design capabilities in large language models, with implications for both AI research and the broader scientific enterprise.

\textbf{Advancing AI-Assisted Research.} By providing a standardized evaluation framework, \textsc{SCOPE} enables rigorous benchmarking of LLMs on a core scientific reasoning task that has received limited systematic attention. The fine-grained evaluation dimensions---spanning high-level experimental logic and low-level resource configuration---offer actionable diagnostics that can guide targeted model improvements. We view this as a step toward AI systems that can meaningfully assist researchers in designing rigorous, comprehensive experiments, particularly benefiting early-career researchers and those in resource-constrained settings where access to expert experimental design mentorship is limited.

\textbf{Promoting Reproducibility and Rigor.} The benchmark's structured output format encourages models to produce complete, verifiable experimental plans with explicit dataset sources, baseline specifications, and metric definitions. This emphasis on transparency and provenance aligns with broader efforts to improve reproducibility in machine learning research. Moreover, the redline mechanism explicitly penalizes source hallucination and metric incompatibility---failure modes that, if undetected, could directly undermine the credibility of AI-assisted experimental designs.

\textbf{Workflow Transparency.} The \textsc{SCOPE} evaluation framework is designed with transparency as a first-order requirement: every score is accompanied by a dimension-level justification, the redline mechanism identifies specific fatal flaws, and the full provenance trail of each task is preserved. This design choice reflects our view that AI-assisted experimental design tools should operate as auditable reasoning partners rather than opaque recommendation engines. The structured decomposition of experimental design into independently evaluable sub-dimensions provides a natural interface for human researchers to review, challenge, and refine AI-generated experimental plans before committing resources to execution.

\textbf{Potential Risks and Mitigations.} We acknowledge two considerations regarding the broader deployment of systems like \textsc{SCOPE}. First, over-reliance on AI-generated experimental plans without adequate human oversight could lead to a homogenization of experimental protocols, where standard benchmark-task-metric combinations are favored at the expense of creative or unconventional validation strategies. Mitigating this risk requires that AI-assisted experimental design tools remain subordinate to researcher judgment, with the structured, auditable output format of \textsc{SCOPE} facilitating critical review. Second, as models improve on benchmarks like \textsc{SCOPE}, there is a risk of benchmark overfitting, where improvements reflect optimization against the specific evaluation rubric rather than genuine advances in experimental reasoning. Periodic benchmark updates, task rotation, and the development of complementary evaluation modalities are necessary to maintain the benchmark's validity as a measure of progress. We emphasize that \textsc{SCOPE} is designed as a research instrument for understanding and improving LLM capabilities, not as a production system for autonomous experimental design, and that human expertise remains essential throughout the research process.

\end{document}